\documentclass{article}

\usepackage[table]{xcolor}
\usepackage{iclr2027_conference,times}

\usepackage[hidelinks]{hyperref}
\hypersetup{
  pdftitle={Generalizable VLA Finetuning via Representation Anchoring and Language-Action Alignment},
  pdfauthor={Dwip Dalal, Shivansh Patel, Chahit Jain, Jeonghwan Kim, Utkarsh Mishra, Yuchen Song, Hyeonjeong Ha, Alex Baratian, Heng Ji, Svetlana Lazebnik, Unnat Jain}
}
\usepackage{url}

\iclrfinalcopy

\usepackage{eccvabbrv}     % \eg, \ie, \etc, \cf, \etal, etc.
\usepackage{float}
\usepackage{graphicx}
\usepackage{booktabs}
\usepackage{tabularx}
\usepackage{multirow}
\usepackage{amsmath}
\usepackage{amssymb}
\usepackage{bbm}
\usepackage{enumitem}
\usepackage{adjustbox}
\usepackage{makecell}
\usepackage{todonotes}
\usepackage{pifont}
\usepackage{symbol}
\usepackage{wrapfig}
\usepackage{needspace}
\usepackage{pgfplots}
\pgfplotsset{compat=1.18}
\usepackage{xparse}
\usepackage{capt-of}
\usepackage{subcaption}
\DeclareCaptionFont{subeight}{\fontsize{8}{9.5}\selectfont}
\usepackage{cleveref}
\crefname{figure}{Fig.}{Figs.}
\Crefname{figure}{Fig.}{Figs.}
\crefname{table}{Tab.}{Tabs.}
\Crefname{table}{Tab.}{Tabs.}
\crefname{section}{Sec.}{Secs.}
\Crefname{section}{Sec.}{Secs.}
\crefname{equation}{Eq.}{Eqs.}
\Crefname{equation}{Eq.}{Eqs.}
\usepackage{xspace}

\newcommand{\secref}[1]{Sec.~\ref{#1}}
\newcommand{\appref}[1]{App.~\ref{#1}}

\definecolor{LightGreen}{rgb}{0.88,1,0.88}
\definecolor{LightRed}{rgb}{1,0.92,0.92}
\definecolor{LightBlue}{rgb}{0.85,0.92,1.0}
\definecolor{LightYellow}{rgb}{1.0,0.97,0.80}
\definecolor{LightPurple}{rgb}{0.93,0.88,1.0}
\definecolor{baseorange}{RGB}{230,126,34}
\definecolor{oursblue}{RGB}{46,116,181}
\definecolor{oursgreen}{RGB}{44,140,44}
\definecolor{mupink}{RGB}{204,51,153}
\definecolor{LightGray}{rgb}{0.92,0.92,0.92}
\definecolor{taskLayout}{RGB}{242,247,230}
\definecolor{taskSpatial}{RGB}{218,218,242}
\definecolor{taskClutter}{RGB}{209,236,248}
\definecolor{taskSemantic}{RGB}{246,228,216}

\title{Generalizable VLA Finetuning via Representation Anchoring and Language-Action Alignment}

\newcommand{\inst}[1]{\textsuperscript{#1}}
\author{%
\textbf{Dwip Dalal}\inst{1},
\textbf{Shivansh Patel}\inst{1},
\textbf{Chahit Jain}\inst{1},
\textbf{Jeonghwan Kim}\inst{1}, \\
\textbf{Utkarsh Mishra}\inst{2},
\textbf{Yuchen Song}\inst{3},
\textbf{Hyeonjeong Ha}\inst{1},
\textbf{Alex Baratian}\inst{1}, \\
\textbf{Heng Ji}\inst{1},
\textbf{Svetlana Lazebnik}\inst{1}\textsuperscript{*},
\textbf{Unnat Jain}\inst{3}\thanks{Equal advising}\\[4pt]
\normalfont\inst{1}University of Illinois Urbana-Champaign \quad
\normalfont\inst{2}Texas A\&M University \\
\normalfont\inst{3}University of California, Irvine}

\begin{document}
\maketitle
% A neutral preprint header, with page numbers retained for the appendix contents.
\fancyhead{}
\pagestyle{plain}
% Keep natural inter-paragraph spacing instead of stretching pages vertically.
\raggedbottom
\thispagestyle{plain}
%===============================================================================
\begin{abstract}

Finetuning a pretrained vision-language model (VLM) on robot demonstrations via behavior cloning (BC) has become the standard recipe for vision-language-action (VLA) policies. However, BC finetuning progressively overwrites the pretrained representations that support visual and semantic generalization. Co-training on web image-text data, a common remedy, applies language and action losses to separate observations, leaving VLAs with language-action misalignment that standard manipulation benchmarks do not expose. We propose \ourmethod, which augments BC with two objectives: \textit{Vision-Language Anchoring} distills layer-wise representations from a frozen VLM copy to prevent this drift, while \textit{Language-Action Alignment} converts each action target into a discrete motion-direction label and jointly trains language and action prediction on the same robot observation. We conduct real-world evaluations across eight manipulation settings on single-arm xArm7 and bimanual YAM robots, using two VLA architectures with regression and flow-matching action heads. Across these settings, \ourmethod consistently improves over BC on novel targets, layouts, and motion-sensitive bimanual tasks requiring coordinated control. At scale in simulation, we demonstrate consistent improvements on OOD perturbations, perceptual robustness, and long-horizon control across LIBERO-PRO, LIBERO-Plus, and CALVIN, respectively, suggesting that preserving pretrained representations and effective action learning are not fundamentally at odds.
Project page: \url{https://anchoralignvla.github.io/}

% Project page: \url{anchoralignvla.github.io}

% Project page: \url{https://anchorvla2026.github.io/AnchorVla2026/}

\end{abstract}

% L2A

% Preserve the manuscript keywords as PDF metadata.
\hypersetup{pdfkeywords={Vision-Language-Action Models, Robot Manipulation, Catastrophic Forgetting, Language-Action Alignment, Out-of-Distribution Generalization}}

\section{Introduction}
\label{sec:intro}
% \vspace{-pt}

Vision-Language-Action (VLA) models adapt pretrained vision-language models (VLMs) into robot manipulation policies by finetuning on expert demonstrations~\citep{zitkovich2023rt2,kim2024openvla}. Action learning typically uses behavior cloning (BC), with either direct regression~\citep{kim2025openvlaoft} or flow matching and diffusion~\citep{black2024pi0}. The premise is that such finetuning should transfer the VLM's semantic priors (understanding of spatial layout, directions, color, shape, etc.) to the resulting control policy. Realizing this promise requires more than retaining pretrained knowledge: finetuning must make it useful for control. 

% We therefore ask: \emph{how can a VLA acquire action-relevant representations while retaining the pretrained priors of VLM that support generalization?}

\begin{figure}[t]
    \centering
    \includegraphics[width=\linewidth]{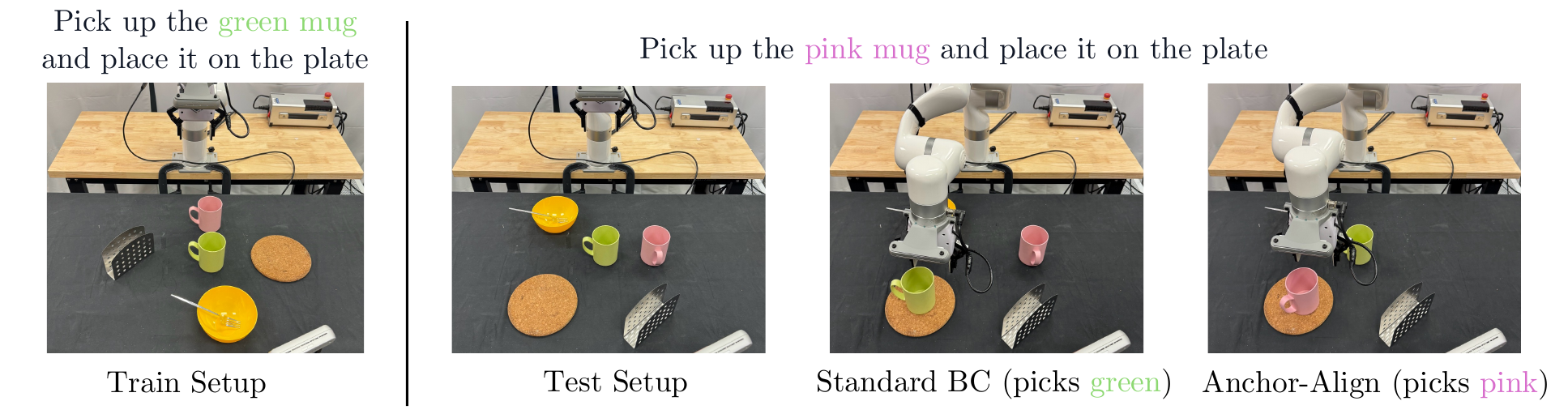}
    \vspace{-20pt}
    \caption{\small \textbf{Out-of-distribution test for VLAs.} Standard BC, finetuned to pick up the green mug, reaches for the green mug even when instructed to pick up the pink mug, whereas \ourmethod retains the VLM's representations and successfully completes the task.}
    \label{fig:ood_figure}
    \includegraphics[width=\linewidth]{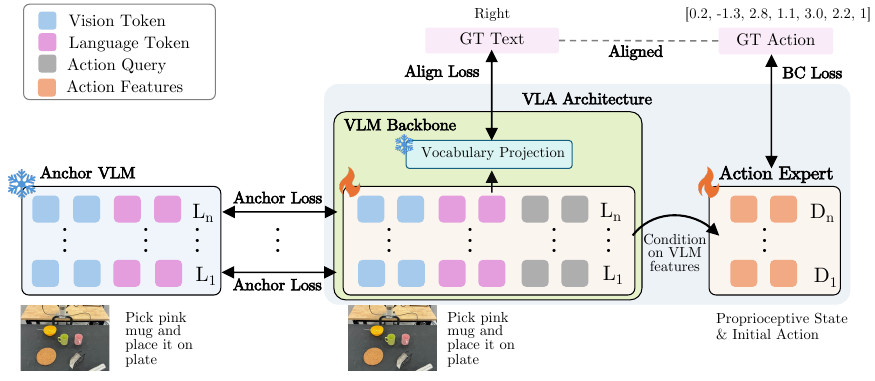}
    \caption{\small \textbf{\small \ourmethod preserves VLM priors and aligns them with action during VLA finetuning.} Vision-Language Anchoring distills representations from a frozen pretrained VLM into the trainable VLA at every transformer layer (Anchor loss), preventing catastrophic forgetting. Language-Action Alignment programmatically converts each ground-truth demonstration trajectory into a discrete language label and trains the model to predict this label on the same observation it acts on (Align loss). Together, these objectives retain the pretrained representations while grounding action generation in VLM semantics.}
    \label{fig:alignment_aware_training}
    \vspace{-10pt}
\end{figure}

Consider a VLA policy finetuned to ``pick up the green mug and place it on the plate'' in a scene containing both a green and pink mug (Fig.~\ref{fig:ood_figure}). Since the two mugs share the same shape and manipulation affordances, the policy should generalize to ``pick up the pink mug'' if finetuning preserves the VLM's color understanding. However, with existing methods, it does not. In fact, BC finetuning corrupts the very prior that makes VLMs worth adapting, leading to two failure modes that persist even under good training practices.

First, standard BC optimizes only the action prediction loss, with nothing protecting the VLM's pretrained representations from being overwritten.
Over the course of finetuning, these updates progressively erase the visuolinguistic and spatial concepts the VLM acquired during internet-scale pretraining.
Concretely, on our physical xArm7 robot, a VLA trained to pick up the green mug reaches for the green mug $90\%$ of the time even when instructed to pick up the pink one, indicating that color grounding has been erased from the backbone.
On LIBERO-PRO's position-swap test, where object positions are rearranged at evaluation, the policy almost always replays its memorized training trajectory rather than acting on the current observation.
To fix this failure mode, recent work has proposed co-training, or augmenting continuous action data with general-purpose VQA, captioning, and scene-description data~\citep{lee2025molmoact,driess2025knowledge,yang2025magma,zhou2025chatvla} to prevent catastrophic forgetting. Yet we find that on the position-swap test, co-trained VLAs score $0\%$. Evidently, losses on such auxiliary data do not transfer the VLM's semantic understanding to the finetuned VLA.
As a more direct remedy, we propose
\textit{Vision-Language Anchoring}, or maintaining a frozen copy of the pretrained VLM (Fig.~\ref{fig:alignment_aware_training}, left) and applying a distillation loss throughout finetuning. This requires no additional data or architectural changes, yet recovers out-of-distribution (OOD) generalization.

The second failure mode runs even deeper. Co-training gives the VLA's two heads disjoint supervision: the action head is supervised by expert demonstrations while the language head is supervised by losses on generic image-text tasks. Because the heads share a backbone but are never supervised on the same robot observation, their predictions may contradict each other: when the action head predicts a movement to the right, the language head may predict ``left.'' To fix this misalignment and further improve task success, we introduce \textit{Language-Action Alignment}, which co-supervises both heads on the same robot observation during finetuning (Fig.~\ref{fig:alignment_aware_training}, top). To supply the aligned supervisory data, we automatically convert continuous action targets in expert demonstrations into discrete action directions expressed in language.

\noindent We propose combining Vision-Language Anchoring and Language-Action Alignment with the standard BC loss to form \textbf{\ourmethod}, a VLA finetuning method with consistent improvements across simulation and real-world experiments under both regression and flow-matching action heads.
In simulation, our method improves OOD generalization on LIBERO-PRO and LIBERO-Plus benchmarks, which introduce unseen spatial rearrangements, camera perturbations, and other shifts absent from training, as well as long-horizon control on CALVIN.
In real-world experiments on an xArm7 robot, \ourmethod improves spatial generalization, target generalization, and target grounding in clutter.
On a bimanual YAM robot, it further improves motion-sensitive control, spanning reactive grasping and coordinated bimanual transport under visual shift.

\vspace{-1pt}

% \vspace{-5pt}

\section{Related Works}
\label{sec:related_works}

\noindent\textbf{General-Purpose VLAs for Robot Manipulation.} VLM-based policies leverage pretrained vision-language representations for instruction-conditioned control~\citep{driess2023palm,li2023vision,qu2025spatialvla,kim2025openvlaoft,zitkovich2023rt2}. Recent general-purpose VLAs pair these pretrained backbones with expressive action-generation modules, including flow-matching action experts~\citep{black2024pi0,black2025pi05} and diffusion transformers~\citep{bjorck2025grootn1}.
VLA-Adapter~\citep{wang2025vla} enables direct adaptation without large-scale robotic pretraining through lightweight Bridge Attention, while StarVLA~\citep{starvla2026} provides a modular framework for combining backbones and action heads.
We build on these two frameworks, finetuning pretrained VLMs directly on expert demonstrations.

\noindent\textbf{Preserving VLM Representations in VLAs.}
Pretrained visual representations are a well-established driver of robot manipulation performance~\citep{nair2022r3m,radosavovic2023mvp,dasari2023data4robotics,wang2024hpt}. In VLAs, this requirement extends beyond the vision encoder to the full VLM backbone, since semantic and spatial reasoning are distributed across both vision and language representations.
Prior VLAs commonly mitigate catastrophic forgetting by co-training on robot and generic vision-language data~\citep{yang2025magma,lee2025molmoact,qu2025eo,wen2025dexvla}, with variants adding action-grounding objectives, perception tokens, or chain-of-thought plans ~\citep{yang2025magma,lee2025molmoact,zawalski2024robotic,zhou2025chatvla,zhou2025vision}.~\citet{grover2025enhancing} instead preserves pretrained features architecturally, anchoring a frozen vision encoder with a trainable one. MAPS~\citep{huang2025maps} penalizes each module's weights for drifting from the pretrained checkpoint, constraining vision layers more tightly than action-oriented ones, and \citet{kachaev2025don} aligns visual patch features at a single intermediate layer with a frozen vision foundation model. Anchor-Align instead applies two simple losses at the VLM backbone level throughout finetuning, jointly preserving the pretrained VLM prior and enriching its representations with language supervision grounded in continuous robot actions. This couple's preservation with action-relevant adaptation, rather than treating preservation as an end goal.

\noindent\textbf{Language supervision for action learning.}
Language supports action learning through intermediate predictions,
action representations, and auxiliary supervision.
Pretrained VLM knowledge has shown great support for planning and action selection in visual navigation~\citep{dalal-etal-2026-mllms}.
RT-H~\citep{belkhale2024rt} conditions low-level actions on predicted
language motions, while ECoT~\citep{zawalski2024robotic} and
InSpire~\citep{zhang2025inspire} introduce grounded reasoning chains
and object--gripper spatial relations as textual intermediates.
VLA-0~\citep{goyal2025vla0} serializes discretized actions as text,
whereas Actions-as-Language~\citep{hancock2025actions} expresses
low-level commands in natural language within a reasoning hierarchy.
LAP~\citep{zha2026lap} supervises structured action-chunk descriptions;
its LAP-3B implementation pairs this objective with a continuous-action
expert whose gradients are blocked from the VLM.
Anchor-Align instead preserves pretrained representations while adapting
the VLM backbone through continuous-action gradients.
Vision-Language Anchoring constrains representational drift, while
Language-Action Alignment grounds language predictions in the same
action targets supervised by BC.
Language thus supplements continuous-action learning rather than
replacing it, with both objectives using only existing demonstrations
and requiring no intermediate language generation at inference.

\vspace{-1pt}
\vspace{-5pt}

\section{\ourmethod Method}
\label{sec:method}
\vspace{-5pt}

This section presents \ourmethod~(Fig.~\ref{fig:alignment_aware_training}), a recipe that finetunes a pretrained VLM into a continuous action policy while preserving its pretrained VL representations. We first describe our base VLA architecture and training objective (Sec.~\ref{sec:base_vla}), then the two objectives \ourmethod adds on top of standard BC: \textit{Vision-Language Anchoring} (Sec.~\ref{sec:anchoring-vlm}) and \textit{Language-Action Alignment} (Sec.~\ref{sec:lang-action-align}).

% For the action expert, it sgcan be either flow-matching action sion and flow-matching action heads

% \vspace{-0.15cm}

% \vspace{-0.15cm}
% \vspace{-3pt}

\subsection{Base VLA Architecture}
\label{sec:base_vla}

% \vspace{-3pt}

We consider VLAs that adopt a pretrained vision-language backbone with a dedicated action head \citep{starvla2026, wang2025vla}, as shown in Figure \ref{fig:alignment_aware_training}. The backbone encodes the visual observation and language instruction into a hierarchy of multimodal representations. Separately, robot proprioception is encoded by a lightweight state projector and provided to the action pathway as a conditioning signal. A fixed set of learned action queries extracts task-relevant information from the vision-language representations, either through joint self-attention within the backbone or through cross-attention in the action head. The action head then combines the resulting query features with the encoded robot state to predict a fixed-horizon chunk of continuous 7-DoF actions (3D translation and rotation offsets plus gripper control) using either direct regression or conditional flow matching.

We train this architecture with three objectives. The primary term is the standard BC loss $\mathcal{L}_{\text{action}}$ between the predicted and ground-truth action chunks, $\hat{\mathbf{a}}^{\text{cont}}$ and $\mathbf{a}^{\text{cont}}$ respectively. The form of the BC loss depends on the action-head architecture: an $L_1$ loss for a regression action head, and an $L_2$ loss over the predicted velocity for a flow-matching action head. To it we add two objectives, both evaluated on the same robot observation: an anchoring loss $\mathcal{L}_{\text{anchor}}$ that preserves the backbone's pretrained vision-language representations (Sec.~\ref{sec:anchoring-vlm}), and an alignment loss $\mathcal{L}_{\text{align}}$ that grounds the model's language predictions in its actions (Sec.~\ref{sec:lang-action-align}). The three terms are optimized jointly:
\vspace{-2pt}
\begin{equation}
\label{eq:total_loss}
\mathcal{L}_{\text{total}} = \mathcal{L}_{\text{action}} + \lambda_{\text{anchor}}\,\mathcal{L}_{\text{anchor}} + \lambda_{\text{align}}\,\mathcal{L}_{\text{align}},
\end{equation}
\vspace{-3pt}
where $\lambda_{\text{anchor}}$ and $\lambda_{\text{align}}$ weight the two added objectives against the action loss.

% \lana{What is the standard BC loss? Is it CE loss? Be more explicit.}
% \vspace{-3pt}

\subsection{Vision-Language Anchoring}
\label{sec:anchoring-vlm}
% \vspace{-5pt}
% \vspace{-3pt}

To prevent catastrophic forgetting, we anchor the VLA's backbone VLM with a frozen copy of the same VLM (the \textit{anchor} VLM), which processes the same input batch (images, text) in parallel. This distills the original pretrained VLM's representations into the trainable backbone. In effect, the anchor VLM serves as a frozen teacher, a role well established in the ML literature: feature distillation trains a student to match a teacher's representations~\citep{hinton2015distilling, romero2015fitnets, jiao2020tinybert}, and anti-forgetting methods regularize a finetuned model toward its pretrained copy in weight space~\citep{kirkpatrick2017overcoming, xuhong2018explicit} or representation space~\citep{li2017learning, douillard2020podnet, mukhoti2024fine}. We anchor hidden states of the vision and text tokens between the backbone and anchor VLM at every decoder layer $u \in \mathcal{D}$, where $\mathcal{D}$ is the set of decoder layers. Let $\mathbf{m}$ denote the set of vision and text positions. The anchoring loss is the mean squared error (MSE) between the two sets of hidden states, averaged over all $|\mathcal{D}|$ decoder layers:
\vspace{-5pt}
\begin{equation}
\mathcal{L}_{\text{anchor}} = \frac{1}{|\mathcal{D}|}\sum_{u\in\mathcal{D}}\frac{1}{N d}\sum_{i\in\mathbf{m}}\sum_{j=1}^{d}\big( \mathbf{H}^{S}_{u}[i,j] - \mathbf{H}^{A}_{u}[i,j] \big)^2,
\end{equation}
\vspace{-5pt}
\vspace{-2pt}
\vspace{-2pt}

where $\mathbf{H}^{S}_{u}$ and $\mathbf{H}^{A}_{u}$ are the hidden states of the trainable backbone ($S$) and the anchor ($A$) at decoder layer $u$, $\mathbf{H}_{u}[i,j]$ is the $j$-th feature of the token at position $i$, $N=|\mathbf{m}|$ is the number of vision and text tokens, and $d$ is the hidden dimension of the VLM.

\vspace{-0.15cm}

% \begin{equation}
%     z_i = W_{\mathrm{proj}} h_i, \qquad
%     \ell_i = W_{\mathrm{lm}} z_i, \qquad
%     \mathcal{L}_{\mathrm{align}} = \mathrm{CE}(\ell_i, y_i).
% \end{equation}

% Given the image and instruction, the model should be able to
% predict what direction the robot is about to move, expressed as a real English word.
% Requiring this prediction to be natural language forces the pre-action
% representation to remain language-grounded while remaining informative for
% downstream action decoding. Supervising this position also sends gradients through
% both the vision and language pathways without directly manipulating vision patch
% embeddings.

% To tie such language representations to the physical action space, we supervise a single
% token that sits at the exact boundary where language ends and action generation begins.  The hidden state of the last text token, immediately preceding the first action token,
% satisfies both requirements via causal self-attention: it attends to all vision patches
% and all text tokens, making it a natural summary of everything the model has observed
% before producing an action.

\begin{figure}
    \centering
    \includegraphics[width=0.85\linewidth]{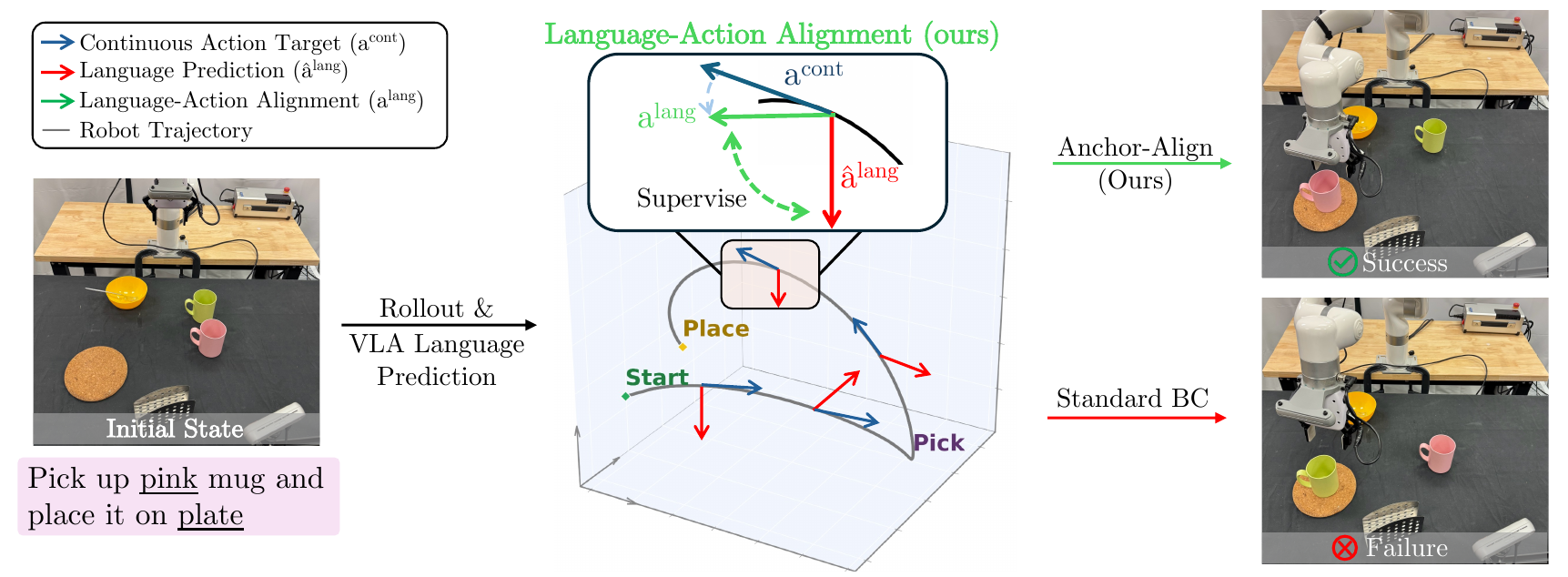}
    \caption{\small \textbf{\ours aligns the VLM's language predictions with action generation during VLA finetuning.} Standard BC supervises only action prediction, allowing the model's language output (red) to contradict the required motion on the same observation (blue). Language-Action Alignment derives a discrete motion-direction label from each ground-truth action target and trains the backbone to predict it on the same observation (green), bringing language and action into agreement.}
    \label{fig:teaser}
    \vspace{-10pt}
\end{figure}

\subsection{Language-Action Alignment}
\label{sec:lang-action-align}
% \vspace{-0.15cm}
% \vspace{-5pt}
% \vspace{-3pt}

As illustrated in Fig.~\ref{fig:teaser}, we programmatically convert the continuous action target $\mathbf{a}^{\text{cont}}$ into a motion-direction label $\mathbf{a}^{\text{lang}}$ for each non-stationary demonstration chunk and use it to supervise the VLM backbone on the same observation used for continuous action prediction. We use the last-layer hidden state of the last instruction token, immediately before action prediction, which summarizes the visual prefix and language instruction. With $N$ vision and text tokens, this pre-action hidden state is
$\mathbf{h}^{\text{pre}} = \mathbf{H}^{S}[N] \in \mathbb{R}^{d}$,
where $\mathbf{H}^{S}$ denotes the backbone VLM's last-layer hidden states. We project $\mathbf{h}^{\text{pre}}$ onto the vocabulary through a learned projection and the frozen pretrained language head:
$\hat{\mathbf{a}}^{\text{lang}} =
\mathbf{W}_{\text{lm}}\mathbf{W}_{\text{proj}}\mathbf{h}^{\text{pre}}
\in \mathbb{R}^{|\mathcal{V}|},$
where $\mathbf{W}_{\text{proj}} \in \mathbb{R}^{d \times d}$ is learned and
$\mathbf{W}_{\text{lm}} \in \mathbb{R}^{|\mathcal{V}| \times d}$ is the frozen pretrained LM head. The alignment loss is the cross-entropy between these vocabulary logits and the ground-truth direction label:
\vspace{-5pt}
\begin{equation}
\mathcal{L}_{\text{align}} =
\mathrm{CE}(\hat{\mathbf{a}}^{\text{lang}},\, \mathbf{a}^{\text{lang}}).
\end{equation}
\vspace{-5pt}
\vspace{-5pt}
\vspace{-5pt}

\vspace{-5pt}

\textbf{Language Label Construction.} The alignment targets are derived from ground-truth action trajectories with no human annotation. For each training sample, we assign one of six interpretable direction labels,
$W=\{\emph{up}, \emph{down}, \emph{left}, \emph{right}, \emph{forward}, \emph{backward}\}$,
through three steps: average chunking, filtering, and discretization. Given the batch of action targets
$\mathbf{A}\in\mathbb{R}^{B\times K\times 7}$ ($\mathbf{a}^{\text{cont}}$ stacked over $B$ batch samples, each a $K$-step chunk), we average the translational components across the chunk dimension to reduce action noise:
$\bar{\mathbf{v}}=\frac{1}{K}\sum_{k=1}^{K}\mathbf{A}_{:,k,1:3}\in\mathbb{R}^{B\times 3}$. We then remove near-stationary samples satisfying
$\|\bar{\mathbf{v}}_i\|_2<\tau$ (where $i$ indexes samples in the batch), with $\tau$ a small stationarity threshold. For each remaining sample, we choose the dominant translation axis,
$j^\star=\arg\max_{j\in\{x,y,z\}}|\bar{v}_{i,j}|$,
where $\bar{v}_{i,j}$ is the $j$-th scalar component of $\bar{\mathbf{v}}_i$, and use the sign of $\bar{v}_{i,j^\star}$ to select the corresponding direction word. The resulting direction label $\mathbf{a}^{\text{lang}}$ supervises the language-alignment loss. Training implementation details are given in \appref{subsec:align_head}; the diagnostic evaluation using these direction labels is described in \appref{benchmark_curation}.

% \vspace{-0.15cm}

\vspace{-5pt}

\section{Experiments}
\vspace{-5pt}
\label{sec:experiments}

We evaluate on simulation benchmarks and real-world robot experiments. \secref{sec:benchmarks} describes the benchmarks, baselines, and implementation details. \secref{sec:quantitative} and \secref{sec:ablation} report results and ablations on LIBERO-PRO, LIBERO-Plus, and CALVIN. \secref{sec:real_world} results on an xArm7 and a bimanual YAM.

% \vspace{-5pt}

% \vspace{-10pt}
\subsection{Experimental Setup}
\label{sec:benchmarks}

\paragraph{Benchmarks.} We evaluate on LIBERO~\citep{liu2023libero}, LIBERO-PRO~\citep{zhou2025liberopro}, LIBERO-Plus~\citep{fei2025liberoplus}, and CALVIN. LIBERO comprises four 10-task suites (Spatial, Object, Goal, Long) whose test setups follow the training distribution, probing only in-distribution performance and now being largely saturated. We therefore report standard LIBERO results in App.~\ref{subsec:standard_libero}, where \ours outperforms all baselines, and focus below on the more challenging LIBERO-PRO and LIBERO-Plus stress tests.
LIBERO-PRO~\citep{zhou2025liberopro} evaluates \emph{semantic generalization} under OOD perturbations: language rephrase, object swap, and position swap. LIBERO-Plus~\citep{fei2025liberoplus} evaluates \emph{perceptual robustness} across seven axes reflecting real-world deployment variability: camera viewpoint, lighting, background texture, object layout, robot initial state, language instruction, and sensor noise. Subcategory details are provided in App.~\ref{app:benchmark_subcategories}. Finally, CALVIN ABC$\rightarrow$D tests long-horizon, language-conditioned manipulation by training on environments A, Bx, and C and evaluating zero-shot on held-out environment D.
We further extend our study by collecting a dataset of more than 300 real-world demonstrations.
We use this dataset to evaluate performance across 8 manipulation tasks.

\vspace{-10pt}

\paragraph{Baselines.}
We compare five baseline families (details in App.~\ref{app:baseline_details}): (i) the VLA models OpenVLA-OFT~\citep{kim2025openvlaoft} and \vlaadapter~\citep{wang2025vla} (our base model); (ii) the representation-preserving baselines \frozen~\citep{wang2025vla}, which freezes the backbone, and MAPS~\citep{huang2025maps}, our reimplementation of module-wise proximity scheduling on \vlaadapter; (iii) the co-trained VLAs: MolmoAct~\citep{lee2025molmoact} and \textit{Co-training + KI}, our reimplementation of Knowledge Insulation~\citep{driess2025knowledge} on \vlaadapter; (iv) the language-based action baselines VLA-0~\citep{goyal2025vla0}, which represents discretized actions as text, and LAP~\citep{zha2026lap}, which supervises language descriptions of action-chunk motion; and (v) the CALVIN baselines UniVLA~\citep{bu2025univla} and OpenHelix~\citep{cui2025openhelix}.

\vspace{-10pt}
\paragraph{Implementation Details.} We use Prismatic-Qwen2.5-0.5B~\citep{yang2024qwen25} as VLM backbone and finetune it with LoRA~\citep{hu2022lora} with rank $r=64$, applied on all layers. The input sequence is vision + text + proprioceptive state. We use a regression action head with bridge attention to layer-wise condition the action head latents on the VLM features~\citep{wang2025vla}. Image encoding, hyperparameters, and other implementation details are provided in App.~\ref{app:implementation_detail}. We additionally evaluate StarVLA~\citep{starvla2026}, which pairs Qwen2.5-VL~\citep{bai2025qwen25vl} with a flow-matching GR00T FM-DiT~\citep{bjorck2025grootn1} action head, in our real-world experiments.

% \lana{Add references to LoRA, DINOv2, and SigLIP}

\begin{table*}[t]
    \centering
    \caption{\textbf{Success rates on the LIBERO-PRO and LIBERO-Plus benchmarks.}
    \ours achieves the highest mean on both benchmarks. The best result in each column is \textbf{bolded} and the second-best is \underline{underlined}. The 5-seed mean and variance are reported in App.~\ref{app:multi_seed}, confirming that these gains are statistically significant.
    }
    \label{tab:libero_suite}
\begin{adjustbox}{max width=\textwidth}
    \setlength{\tabcolsep}{3.5pt}
    \renewcommand{\arraystretch}{1.2}
    \begin{tabular}{@{}l | ccc c | ccccccc c@{}}
    \toprule
    &
    \multicolumn{4}{>{\columncolor{LightBlue}}c|}{\textbf{LIBERO-PRO}} &
    \multicolumn{8}{>{\columncolor{LightRed}}c}{\textbf{LIBERO-Plus}} \\
    \cmidrule(lr){2-5} \cmidrule(lr){6-13}
    \textbf{Method} &
    \makecell{Lang.\\Reph.} & \makecell{Object\\Swap} & \makecell{Pos.\\Swap} & \makecell{Mean} &
    \makecell{Lang.\\Instr.} & \makecell{Bg.\\Text.} & \makecell{Robot\\Init} & \makecell{Cam.\\View} & \makecell{Obj.\\Layout} & \makecell{Light\\Cond.} & \makecell{Sensor\\Noise} & \makecell{Mean} \\
    \midrule
    
    % MolmoAct Plus Mean recomputed with the per-axis episode counts noted at the KI row: 1,439/2,402 = 59.9 (was 60.8).
    \molmoact~\citep{lee2025molmoact}      & 77.8 & 82.4 & 0.0 & 53.4 & 79.5 & 84.1 & 47.4 & 10.1 & 76.5 & 77.4 & 53.4 & 59.9 \\
    VLA-0~\citep{goyal2025vla0} & 75.4 & 93.0 & 5.4 & 57.9 & 60.0 & 70.2 & 34.9 & 42.6 & 83.9 & 88.7 & 60.1 & 62.0 \\
    OpenVLA-OFT~\citep{kim2025openvlaoft}      & 74.4 & \underline{95.2} & 0.0 & 56.5 & 81.5 & 95.7 & 40.3 & 94.7 & 88.6 & 95.5 & 28.2 & 74.1 \\
    \vlaadapter~\citep{wang2025vla}           & 91.1 & 89.6 & 2.3 & 61.0 & 85.1 & 90.7 & 52.6 & 92.6 & \underline{93.2} & 93.2 & 89.5 & 85.1 \\
    VLA-A [Frozen]~\citep{wang2025vla}       & 56.0 & 73.4 & 0.0 & 43.1 & 41.5 & 70.9 & 35.1 & 94.4 & 62.3 & 84.9 & 36.2 & 59.9 \\
    % \arrayrulecolor{black!30}\hline\arrayrulecolor{black}
    % Corrected ki-v2, selected checkpoint 005000, evaluation seed 42 (2026-09-23).
    % PRO: eval-ki-v2-5000-int-v3; 268/500 language, 460/500 object, 0/500 position.
    % Standard Spatial: 473/500. Plus: plus-ki-v2-5000-proven-v4/verified_results.json.
    % Plus Mean is episode-weighted over the 2,402 LIBERO-Spatial tasks (Lang. Instr. 390, Bg. 258, Robot 350,
    % Cam. 376, Layout 385, Light 292, Noise 351; from LIBERO-plus task_classification.json): 1,739/2,402 = 72.4.
    Co-training + KI~\citep{driess2025knowledge}  & 53.6 & 92.0 & 0.0 & 48.5 & 61.5 & 93.8 & 48.6 & 89.6 & 77.1 &  96.9 & 48.4 & 72.4 \\
    % VLA-0: August 5--8 evaluations of the released checkpoint, seed 7,
    % action horizon 1, ensemble size 8, greedy decoding. PRO: 377/500,
    % 465/500, 27/500 successes. Plus: latest saved result per task;
    % sensor noise 13/247, overall 1292/2298 (104 trials unavailable).
    
    LAP~\citep{zha2026lap} & \underline{93.2} & 95.0 & \underline{12.4} & \underline{66.9} & \textbf{88.1} & \underline{99.2} & \underline{54.0} & 72.6 & 92.5 & 95.5 & 76.1 & 81.8 \\
    % MAPS on VLA-Adapter (2026-09-20/21): the authors' SPD optimizer (their spd.py, byte-identical) with
    % their layer-wise proximity schedule, lambda_max = 0.5 (their VLA-Adapter/LIBERO value); everything
    % else is the BC recipe (LoRA r64, bs 32x4, lr 2e-4, 10k steps). 10k checkpoint, eval seed 21.
    % PRO: 465/500, 451/500, 58/500 successes. Plus: 2074/2402 (all 2,402 tasks).
    MAPS~\citep{huang2025maps} & 93.0 & 90.2 & 11.6 & 64.9 & 87.1 & 97.3 & 52.9 & \underline{95.2} & 87.5 & \underline{97.3} & \underline{90.6} & \underline{86.3} \\
    % Kachaev et al. visual representation alignment on VLA-Adapter (2026-09-20/21): frozen C-RADIOv3-L
    % teacher, frozen random MLP projector (dim 2048), cosine loss, coefficient 0.2, one mid-depth LLM layer
    % (12 of 24), both cameras; published configuration, NOT tuned for the L1 action loss. Same BC recipe,
    % 10k checkpoint, eval seed 21. PRO: 186/500, 379/500, 3/500 successes. Plus: 1491/2402 (all 2,402 tasks).
    % Vis.\ Rep.\ Align.$^{\ddagger}$~\citep{kachaev2025don} & 37.2 & 75.8 & 0.6 & 37.9 & 48.7 & 75.2 & 45.1 & 54.3 & 60.0 & 82.5 & 77.8 & 62.1 \\
    \rowcolor{LightGreen}
    % \anchor           & 95.4 & 96.2 & 12.6 & \textbf{90.5} & 96.9 & 56.0 & 95.2 & 95.8 & \textbf{100.0} & 95.7 & 89.7 \\
    \rowcolor{LightGreen}
    \ours & \textbf{97.0} & \textbf{96.2} & \textbf{22.6} & \textbf{71.9} & \underline{87.2} & \textbf{99.6} & \textbf{59.1} & \textbf{96.3} & \textbf{97.4} & \textbf{99.0} & \textbf{96.9} & \textbf{90.3} \\
    \bottomrule
    \end{tabular}
    \end{adjustbox}
    
    \vspace{-10pt}
    \end{table*}

% \vspace{-0.15cm}

\subsection{Simulation Benchmark Comparison}
\label{sec:quantitative}
% \vspace{-5pt}

% In this section, we present the main empirical results of the paper. Additional results (including results on the LIBERO suite) and supporting analyses are provided in the supplementary (App.\ref{app:extra_results}).

% \vspace{-5pt}

\noindent\textbf{\ourmethod improves visual grounding and spatial understanding.}\par

\begin{wrapfigure}{r}{0.6\linewidth}
    \centering
    \vspace{-10pt}
    \includegraphics[width=\linewidth]{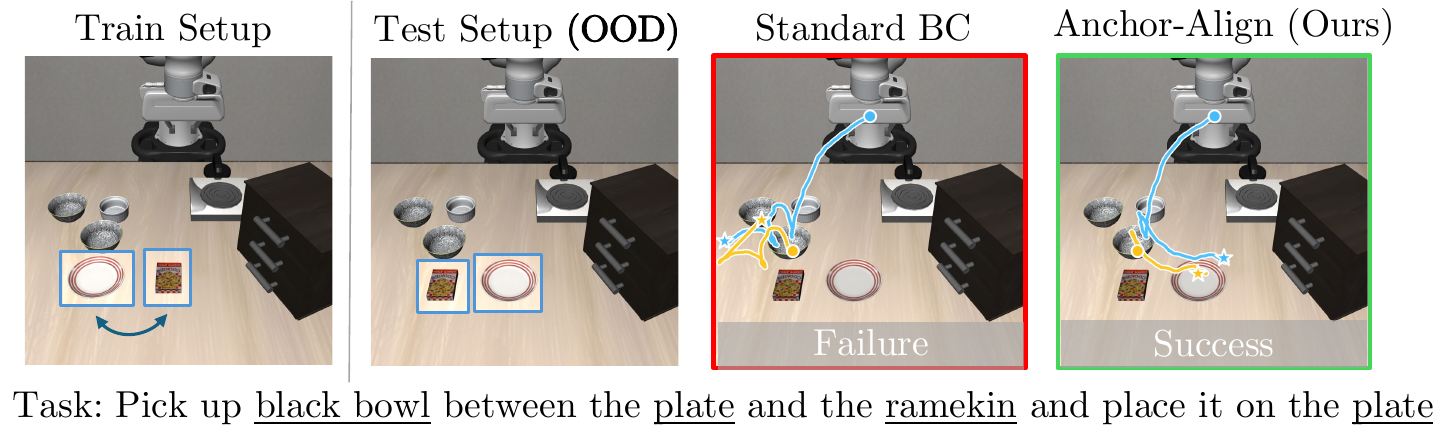}
\captionsetup{font=normalsize,justification=raggedright,singlelinecheck=false,skip=11pt}
    \caption{LIBERO-PRO Swap, performance of \ourmethod in OOD condition.}
    \label{fig:swap_qualitative}
    \vspace{-10pt}
\end{wrapfigure}

Tab.~\ref{tab:libero_suite} shows that \ours achieves the highest mean on both LIBERO-PRO and LIBERO-Plus. For LIBERO-PRO, the position-swap axis is the hardest: task-relevant objects are rearranged, so a policy that memorized fixed scene-to-action mappings fails. Most baselines fail almost entirely under this setting, while even the strongest alternatives reach only 12.4\%; in contrast, \ours achieves 22.6\%. %We attribute this to anchoring preserving the VLM's spatial and object representations, while alignment grounds predictions in the current observation rather than memorized trajectories.
Fig.~\ref{fig:swap_qualitative} shows qualitative examples from LIBERO-PRO, where \ours generalizes to the perturbed scene, while standard BC replays its training trajectory.

\noindent\textbf{\ourmethod improves robustness to perturbations.}
LIBERO-Plus consists of challenging perturbations like changing background texture, adding sensor noise, or unseen initial robot arm configurations. %that mirror real deployment shifts and degrade prior VLAs: background texture rewrites the visual scene, sensor noise corrupts the observation, and robot initial state starts the arm from configurations never seen in training, directly testing generalization rather than memorization.
As we can see from Tab.~\ref{tab:libero_suite}, \ours improves over \vlaadapter~\citep{wang2025vla} on every axis, with the largest gains on the hardest perturbations: background texture ($+8.9$), sensor noise ($+7.4$), robot initial state ($+6.5$), and lighting ($+5.8$). In absolute terms, \ours succeeds $99.6\%$  under background texture changes and $99.0\%$ under lighting changes. Per-suite robustness breakdowns for the LIBERO Long, Object, and Goal suites in \appref{subsec:per_suite_robustness}.

% \vspace{5pt}

% illustrates this: when objects are swapped relative to training, \baseline executes the memorized motor plan and fails, whereas \ours correctly localizes the target in the rearranged scene.

% \begin{figure}[t]
%     \centering
%     \includegraphics[width=0.7\linewidth]{qualitative examples.png}
%    \caption{\small \textbf{Anchor-and-Align generalizes to semantic perturbations.}
% We test OOD scenes with spatial layout changes (top) and object color/size changes (bottom). Standard BC fails by executing trajectories tied to the training scene, such as reaching toward the original object location or missing the resized target. Anchor-and-Align re-grounds the language instruction in the perturbed observation and completes the task.}
%     \label{fig:swap_qualitative}
%     \vspace{-0.6cm}
% \end{figure}

% \newpage

\begin{wraptable}{r}{0.65\columnwidth}
\centering
\setlength{\tabcolsep}{2.5pt}
\renewcommand{\arraystretch}{1.1}
% \vspace{-10pt}
\caption{Long-horizon generalization on CALVIN ABC$\rightarrow$D}
\vspace{-10pt}

\resizebox{\linewidth}{!}{
\begin{tabular}{@{}l | ccccc | c@{}}
\toprule
\textbf{CALVIN ABC$\rightarrow$D} & \cellcolor{LightBlue}\textbf{1/5 ($\uparrow$)} & \cellcolor{LightBlue}\textbf{2/5 ($\uparrow$)} & \cellcolor{LightBlue}\textbf{3/5 ($\uparrow$)} & \cellcolor{LightBlue}\textbf{4/5 ($\uparrow$)} & \cellcolor{LightBlue}\textbf{5/5 ($\uparrow$)} & \cellcolor{LightYellow}\textbf{Len ($\uparrow$)} \\
\midrule
UniVLA~\citep{bu2025univla} & 95.5 & 85.8 & 75.4 & 66.9 & 56.5 & 3.8 \\
OpenVLA-OFT~\citep{kim2025openvlaoft} & 96.3 & 89.1 & 82.4 & 75.8 & 66.5 & 4.1 \\
OpenHelix~\citep{cui2025openhelix} & 97.1 & 91.4 & 82.8 & 72.6 & 64.1 & 4.1 \\
\vlaadapter~\citep{wang2025vla} & \underline{98.3} & \underline{94.0} & \underline{87.5} & \underline{80.0} & \underline{73.1} & \underline{4.3} \\
\rowcolor{LightGreen}
\ours & \textbf{99.1} & \textbf{95.8} & \textbf{90.6} & \textbf{84.7} & \textbf{77.9} & \textbf{4.5} \\
\bottomrule
\end{tabular}
}
\vspace{-10pt}
\label{tab:calvin_abcd}
\end{wraptable}

\textbf{\ourmethod improves long-horizon generalization.} Tab.~\ref{tab:calvin_abcd} shows results on the long-horizon CALVIN ABC$\rightarrow$D benchmark. In each CALVIN rollout, the policy receives a chain of five language instructions and must complete them consecutively. The $k/5$ columns report the fraction of rollouts that complete the first $k$ instructions, and \textbf{Len} denotes the average number of tasks completed consecutively (out of five). Because the chain advances only if each instruction succeeds, small grounding errors compound over the rollout. We therefore compare against the baselines most relevant to this claim: OpenVLA-OFT and UniVLA, large-scale action-pretrained VLAs; OpenHelix, a recent CALVIN-specific VLA; and \vlaadapter trained with standard BC. \ours is the best at every chain length and attains the longest average rollout, with the gain over \vlaadapter widening on the deepest chains where errors compound, from $+0.8$ at one instruction to $+4.8$ at five. The results indicate preserving and aligning pretrained VLM representations improves long-horizon generalization.

% It also beats the strongest action-pretrained baseline, OpenVLA-OFT, raising five-instruction completion from $66.5\%$ to $77.9\%$, indicating that preserving and aligning pretrained VLM representations improves long-horizon generalization without large-scale robot-action pretraining.

% \vspace{-5pt}
\subsection{Impact of Anchoring and Alignment}
\label{sec:ablation}
% \vspace{-5pt}

\begin{wraptable}{r}{0.42\columnwidth}
\vspace{-\baselineskip}
\centering
\scriptsize
\setlength{\tabcolsep}{2pt}
\renewcommand{\arraystretch}{1.0}
\caption{Ablations on LIBERO-PRO and LIBERO-Plus.}
\label{tab:anchoring_alignment}
\begin{tabular*}{\linewidth}{@{\extracolsep{\fill}}llcc@{}}
\toprule
\textbf{Method} & & \textbf{PRO} & \textbf{Plus} \\
\midrule
\vlaadapter & \textit{Standard BC} & 61.0 & 85.1 \\
Align VLA & \textit{Alignment only} & 65.9 & 88.6 \\
\anchor & \textit{Anchoring only} & 68.1 & 87.3 \\
\midrule
\ours & \textit{Full method} & \textbf{71.9} & \textbf{90.3} \\
\bottomrule
\end{tabular*}
\vspace{-10pt}
\end{wraptable}

\paragraph{Both anchoring and alignment are necessary.}
\cref{tab:anchoring_alignment} presents an ablation analysis that isolates the impact of our proposed additional loss terms by comparing the VLA-Adapter baseline with \anchor (only the layer-wise MSE distillation loss $\Lanchor$), Align VLA (only the language-action alignment loss $\Lalign$), and the full \ours. We can see that each of anchoring and alignment independently improve over standard BC on every axis, and combining them in \ours exceeds either alone. Thus, the two play complementary roles: anchoring preserves the representational substrate that alignment leverages.

\begin{wraptable}{r}{0.42\columnwidth}
\vspace{-\baselineskip}
\centering
\scriptsize
\setlength{\tabcolsep}{2pt}
\renewcommand{\arraystretch}{1.0}
\caption{Semantic-label controls on LIBERO-PRO and LIBERO-Plus.}
\label{tab:shuffle_control}
\label{tab:shuffle_control_app}
\begin{tabular*}{\linewidth}{@{\extracolsep{\fill}}llcc@{}}
\toprule
\textbf{Method} & & \textbf{PRO} & \textbf{Plus} \\
\midrule
\vlaadapter & \textit{Standard BC} & 61.0 & 85.1 \\
\midrule
\textit{Shuffle} & \textit{Control} & 61.4 & 84.9 \\
\textit{Scatter} & \textit{Control} & 63.3 & 85.7 \\
\midrule
\ours & \textit{Full method} & \textbf{71.9} & \textbf{90.3} \\
\bottomrule
\end{tabular*}
\vspace{-10pt}
\end{wraptable}

\noindent\textbf{Semantic alignment beyond auxiliary classification.}
We test whether the gains from Language-Action Alignment depend on
the semantic correspondence between demonstrated motion and its
language label, rather than simply supervising direction prediction.
We retain the same training setup, including anchoring and the frozen
language head, but change the target words assigned to the six motion
directions. \emph{Shuffle} applies a fixed permutation of the six direction words
(e.g., forward $\rightarrow$ right), preserving directional vocabulary while breaking its correspondence to the demonstrated motion.
\emph{Scatter} assigns each direction a fixed word unrelated to motion (e.g., forward $\rightarrow$ purple).
Both controls retain a one-to-one mapping between motion directions and target words, requiring the model to distinguish the same directions without semantically matching words to actions.
As shown in \cref{tab:shuffle_control_app}, both perform substantially worse than the semantically matched formulation, indicating that direction prediction alone is insufficient and that the gains depend on the correct correspondence between motion and language (details in App.~\ref{app:alignment_vs_reg}). Linear probes further show improved motion-direction decoding from the backbone (App.~\ref{app:direction_understanding}). Rollout diagnostics show higher language--action agreement for \ourmethod, positively correlated with task success (App.~\ref{reliability_of_language}).

\begin{wrapfigure}{r}{0.30\linewidth}
    \vspace{-14pt}
    \centering
    \captionsetup{skip=2pt}
    \includegraphics[width=\linewidth]{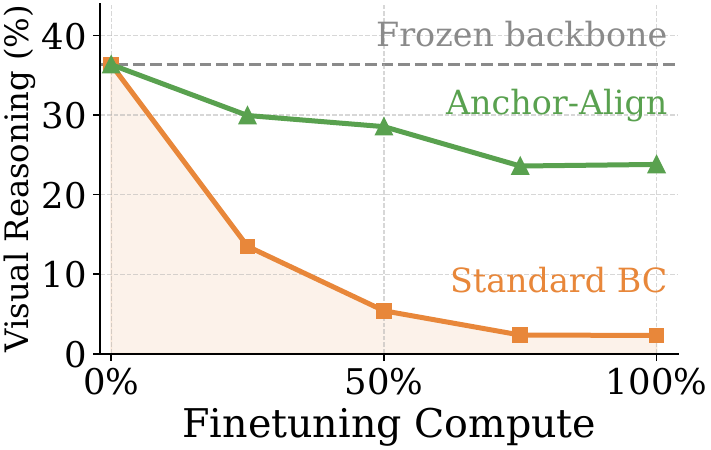}
    \caption{Visual reasoning accuracy during finetuning.}
    \label{fig:catastrophic_forgetting}
    \vspace{-12pt}
\end{wrapfigure}
\noindent \textbf{\ourmethod preserves VLM representations.}
For effective generalization, a VLA must be able to perform visual reasoning, i.e., understand objects, their semantics, and their attributes in a scene, an ability that is a hallmark of pretrained VLM representations. GQA~\citep{hudson2019gqa} tests this ability through compositional questions spanning object, attribute, and spatial understanding. Here, we quantify visual reasoning in standard BC and \ourmethod,  through GQA accuracy over the course of finetuning with the frozen VLM (same VLM used as backbone for VLA model) as an upper bound. As shown in  \cref{fig:catastrophic_forgetting}, standard BC degrades monotonically, losing 94\% of its GQA accuracy within 10K steps (catastrophic forgetting), whereas \ourmethod retains roughly 65\% of the pretrained accuracy. Correspondingly, LIBERO-PRO task success rates (Table~\ref{tab:libero_suite}) are \textcolor{oursgreen}{71.9\%} (\ourmethod), \textcolor{baseorange}{61.0\%} (standard BC), and \textcolor{gray}{43.1\%} (frozen), showing that action performance and vision-language reasoning are not fundamentally at odds. We further substantiate this study by quantifying the change in representation by different finetuning paradigms in App.~\ref{subsec:preservation_decodability}.

\vspace{-5pt}

\subsection{Real-World Results}
\label{sec:real_world}
\vspace{-5pt}

We extend our study to real-world robotic manipulation through controlled experiments across two distinct VLA architectures. We compare \ourmethod against standard BC using \vlaadapter~\citep{wang2025vla}, which pairs a Prismatic-Qwen2.5-0.5B backbone with a Bridge-Attention regression head, and StarVLA~\citep{starvla2026}, which pairs a Qwen2.5-VL backbone with a GR00T FM-DiT flow-matching action head. These models span different VLM scales and action-generation architectures. We evaluate four capabilities: (1) target generalization, (2) spatial generalization, (3) target grounding in clutter, and (4) bimanual coordination.

We evaluate these capabilities across two robotic embodiments. The first three capabilities are evaluated on a single-arm 7-DoF UFactory xArm7, while bimanual coordination is evaluated on the 14-DoF bimanual YAM platform, where the tasks require reactive execution and coordinated motion across both arms. For each evaluated condition, we run 20 rollouts per method and report results as percentages. Each comparison uses the same underlying VLA architecture for \ourmethod and standard BC, differing only in the training objective. Hardware, data collection, and deployment details are provided in App.~\ref{subsec:realworld_impl}, with limitations and error analysis in Sec.~\ref{sec:limitations}.

\begin{figure}[b]
    \vspace{-10pt}
    \centering
    \captionsetup{skip=4pt}
    \captionsetup[subfigure]{skip=2pt}
    \begin{subfigure}[t]{0.49\linewidth}
        \centering
        \includegraphics[width=\linewidth]{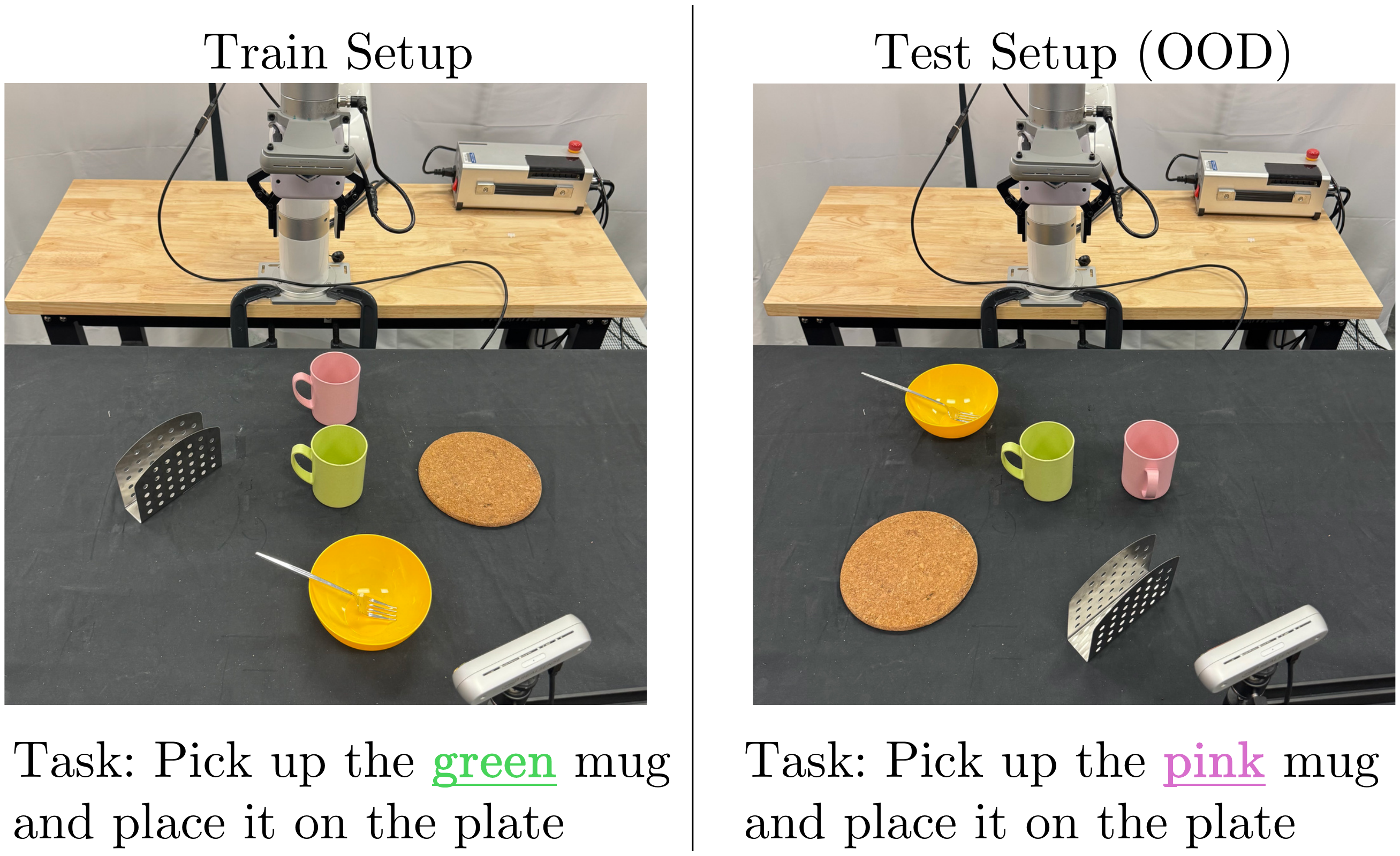}
        \caption{\texttt{Pink Mug Placement}.}
        \label{fig:pink_mug_placement}
    \end{subfigure}\hfill
    \begin{subfigure}[t]{0.49\linewidth}
        \centering
        \includegraphics[width=\linewidth]{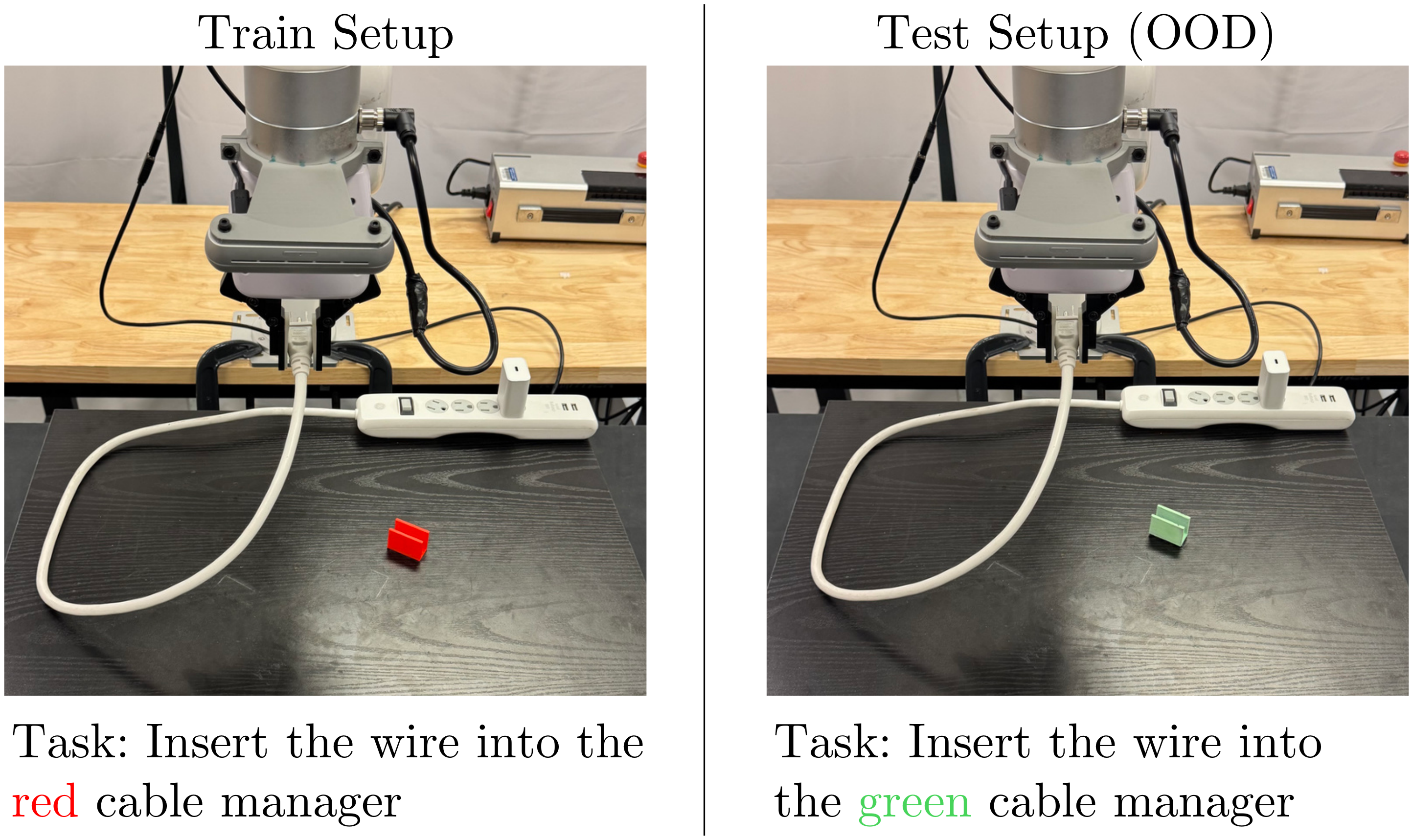}
        \caption{\texttt{Cable Insertion}}
        \label{fig:semantic_generalization_pending}
    \end{subfigure}
    \caption{\small \textbf{Target Generalization (OOD) on xArm7.} \textbf{(a)} Target-object and spatial-layout changes between train \& test.
\textbf{(b)} Target-receptacle color shift changes between train \& test.}
    \label{fig:semantic_generalization}
\end{figure}

\begin{wrapfigure}{r}{0.36\linewidth}
\centering
\vspace{-14pt}
\captionsetup{justification=raggedright,singlelinecheck=false,skip=2pt}
\includegraphics[width=\linewidth]{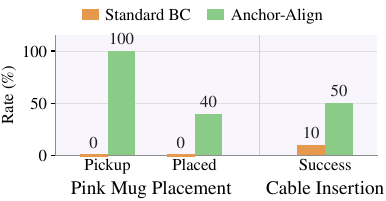}
\caption{Results with StarVLA architecture.}
\label{fig:semantic_results}
\vspace{-13pt}
\end{wrapfigure}

\textbf{Target Generalization (OOD).}
We evaluate target generalization on the xArm7 across two tasks. 1) \texttt{Pink Mug Placement} tests an OOD change in the language-specified object together with an unseen spatial layout. During training, the policy receives the instruction \textit{pick up the green mug and place it on the plate}; at test time, the instructed target changes to the \textit{pink mug} and the objects are presented in a spatial configuration not seen during training (\cref{fig:pink_mug_placement}). Successful execution therefore requires grounding the test-time language to the specified color and object while localizing that target under an unseen layout. 2) \texttt{Cable Insertion} evaluates an OOD change in the language-specified target in a high-precision manipulation task. During training, the policy is instructed to insert a wire into a red cable manager across five spatial configurations. At test time, the instruction specifies a green cable manager, which replaces the red one (\cref{fig:semantic_generalization_pending}). Successful execution requires grounding the new language-specified target and precise control of end-effector position, orientation, and gripper actuation.

With StarVLA, standard BC continues to act on the training target in
\texttt{Pink Mug Placement}: it picks up the green mug in 90\% of trials. In contrast, \ours picks up the instructed
pink mug in 100\% of trials and places it in 40\%. The remaining failures occur during placement, which requires adapting the mug's trajectory to unseen plate locations while maintaining a stable grasp. On \texttt{Cable Insertion}, \ours achieves 50\% success at inserting the wire into
the instructed green cable manager, compared with 10\% for standard
BC (\cref{fig:semantic_results}).

\begin{figure}[t]
\centering
\captionsetup{skip=4pt}
\captionsetup[subfigure]{skip=2pt}
\begin{subfigure}[t]{0.49\linewidth}
    \centering
    % trim 8bp of top margin and add it below: raises (a) ~1pt so its photos line up with (b), same panel height
    \includegraphics[width=\linewidth,trim=0bp -8bp 0bp 8bp,clip]{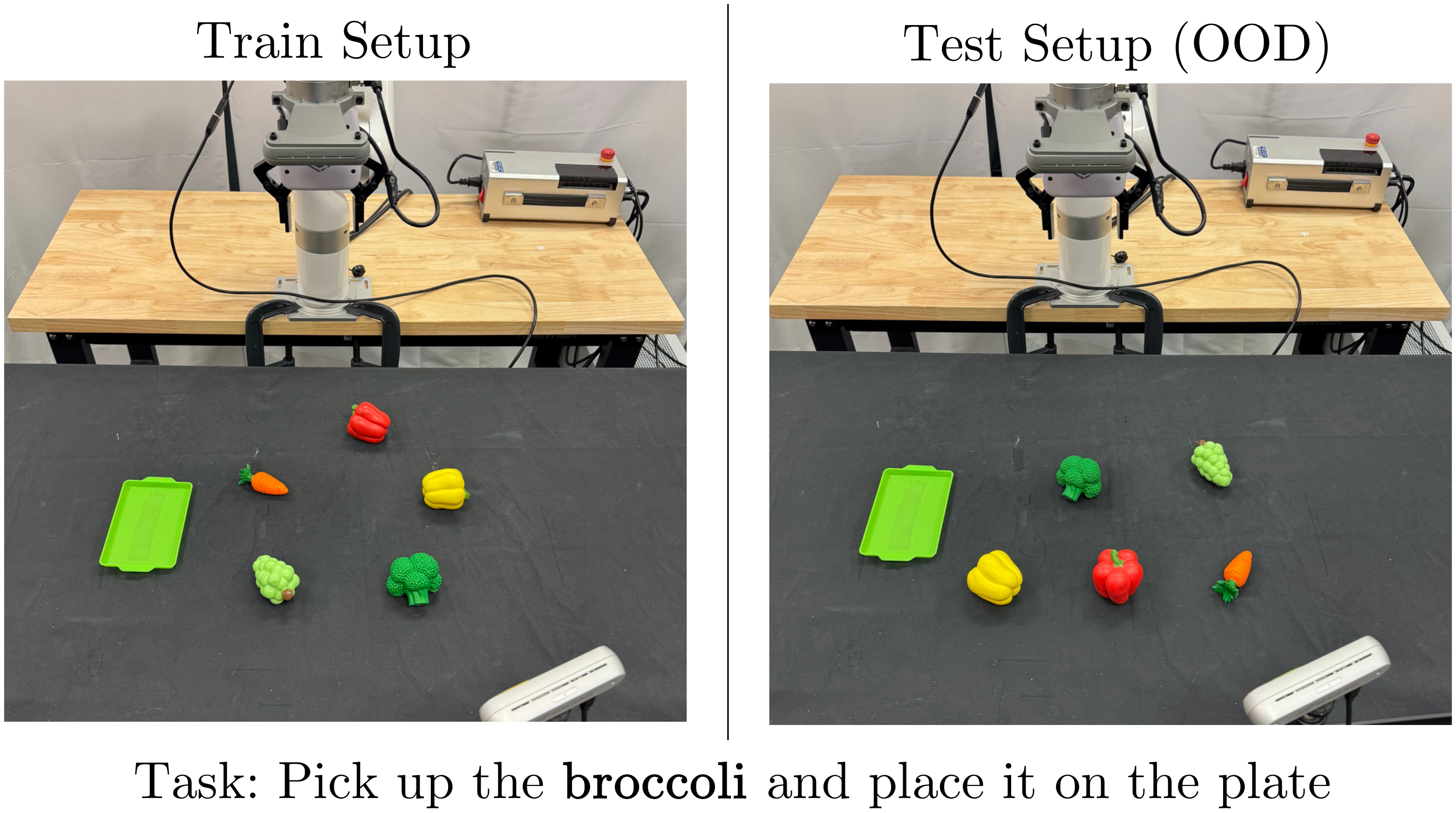}
    \caption{\texttt{Broccoli Placement}.}
    \label{fig:broccoli_placement_ood}
\end{subfigure}\hfill
\begin{subfigure}[t]{0.49\linewidth}
    \centering
    \includegraphics[width=\linewidth]{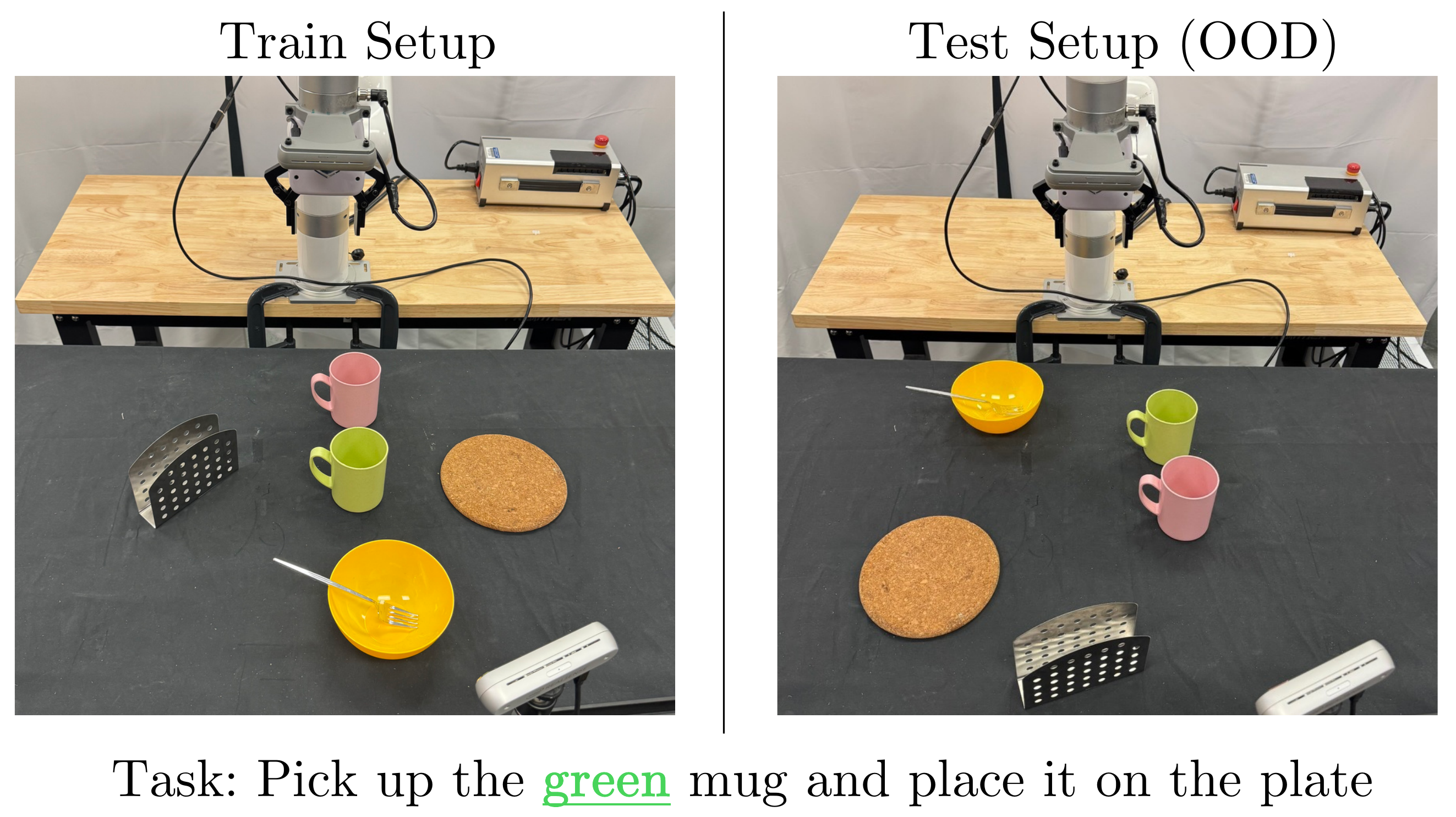}
    \caption{\texttt{Green Mug Placement}.}
    \label{fig:green_mug_placement}
\end{subfigure}
\caption{\textbf{Spatial Generalization (OOD).} \textbf{(a)} \texttt{Broccoli Placement} training and test setups. \textbf{(b)} \texttt{Green Mug Placement} training and test setups.}
\vspace{-10pt}
\label{fig:spatial_grounding_ood}
\end{figure}

\begin{wrapfigure}{r}{0.42\linewidth}
\centering
\captionsetup{justification=raggedright,singlelinecheck=false,skip=4pt}
\vspace{-15pt}
\includegraphics[width=\linewidth]{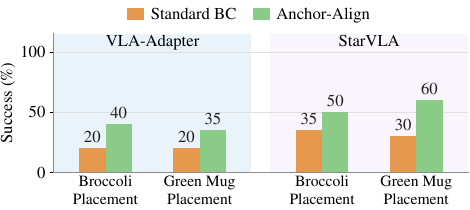}
\caption{Success rates for \texttt{Broccoli Placement} and \texttt{Green Mug Placement}.}
\label{fig:spatial_results}
\vspace{-15pt}
\end{wrapfigure}
\textbf{Spatial Generalization (OOD).}
We evaluate spatial generalization to unseen configurations across two tasks using VLA-Adapter and StarVLA. 1) \texttt{Broccoli Placement} varies the position and orientation of the broccoli and surrounding distractor objects relative to the training configurations (\cref{fig:broccoli_placement_ood}). 2) \texttt{Green Mug Placement} introduces larger layout rearrangements, relocating the target and surrounding objects, including distractors, to configurations unseen during training, similar to the \texttt{Pink Mug Placement}. The instruction and target object remain unchanged: \textit{pick up the green mug and place it on the plate} (\cref{fig:green_mug_placement}).

On \texttt{Broccoli Placement}, \ours improves success from 20\% to 40\% with VLA-Adapter and from 35\% to 50\% with StarVLA. On \texttt{Green Mug Placement}, success increases from 20\% to 35\% with VLA-Adapter and from 30\% to 60\% with StarVLA (\cref{fig:spatial_results}). These improvements across both architectures demonstrate better generalization to unseen spatial configurations while preserving the instruction and manipulation objective within each task.

\begin{wrapfigure}{r}{0.45\linewidth}
\centering
\captionsetup{justification=raggedright,singlelinecheck=false,skip=4pt}
\vspace{-5pt}
\begin{minipage}[c]{0.37\linewidth}
    \centering
    \includegraphics[width=\linewidth]{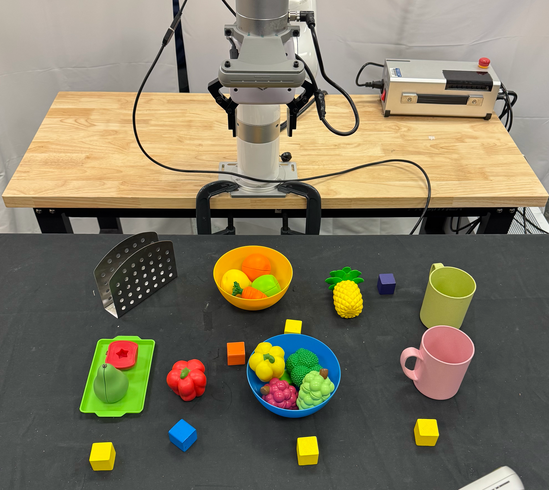}
\end{minipage}
\begin{minipage}[c]{0.55\linewidth}
    \centering
    \includegraphics[width=\linewidth]{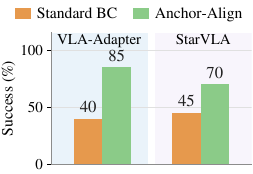}
\end{minipage}
\caption{Left: Real-world evaluation setup. Right: Success rates for \texttt{Clutter Scene.}}
\vspace{-10pt}
\label{fig:realworld_clutter}
\end{wrapfigure}
\textbf{Target Grounding in Clutter.}
We evaluate target grounding in clutter with \texttt{Clutter Scene} using both VLA-Adapter and StarVLA. The robot must pick up the object specified by the language instruction from among multiple distractors. We vary the target across three objects: \texttt{pineapple}, \texttt{yellow pepper}, and \texttt{red pepper}, requiring the policy to ground different instructions within the same scene. On \texttt{Clutter Scene}, \ours increases success from 40\% to 85\% with VLA-Adapter and from 45\% to 70\% with StarVLA (\cref{fig:realworld_clutter}). These gains show that \ours more reliably identifies and manipulates the language-specified target despite surrounding distractors.

% keep Fig. 11 off this page's top so it opens the page that holds its paragraph
\suppressfloats[t]
\begin{figure}[t]
\centering
\captionsetup[subfigure]{font=subeight,justification=centering,singlelinecheck=false,skip=2pt}
% One unbreakable row; equal illustration slots align the subcaptions.
\edef\yamPanelHeight{\the\dimexpr0.222\linewidth\relax}
\makebox[\linewidth][s]{%
\begin{subfigure}[t]{0.194\linewidth}
    \centering
    \begin{minipage}[c][\yamPanelHeight][c]{\linewidth}
    % reactive_grasp.png is pre-cropped to the same 1.15 aspect as the trimmed plate photos, so no trim here.
    \includegraphics[width=\linewidth]{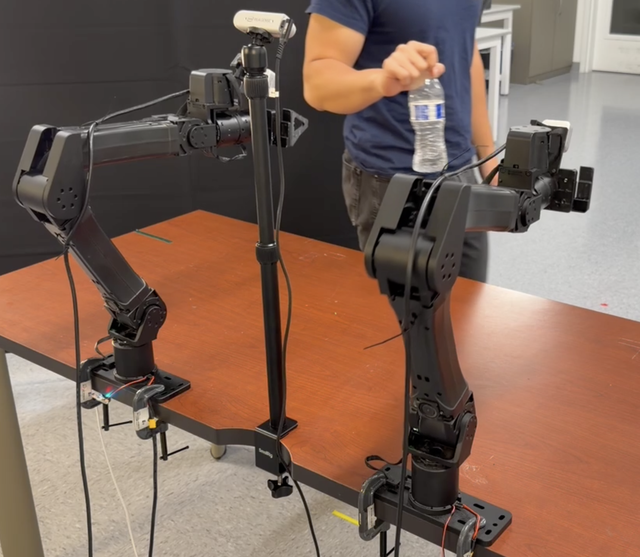}
    \end{minipage}
    \caption{\texttt{Reactive Grasp}.}
    \label{fig:dynamic_grasp}
\end{subfigure}\hfill
\begin{subfigure}[t]{0.194\linewidth}
    \centering
    \begin{minipage}[c][\yamPanelHeight][c]{\linewidth}
    \includegraphics[width=\linewidth,trim=120bp 0bp 434.4bp 0bp,clip]{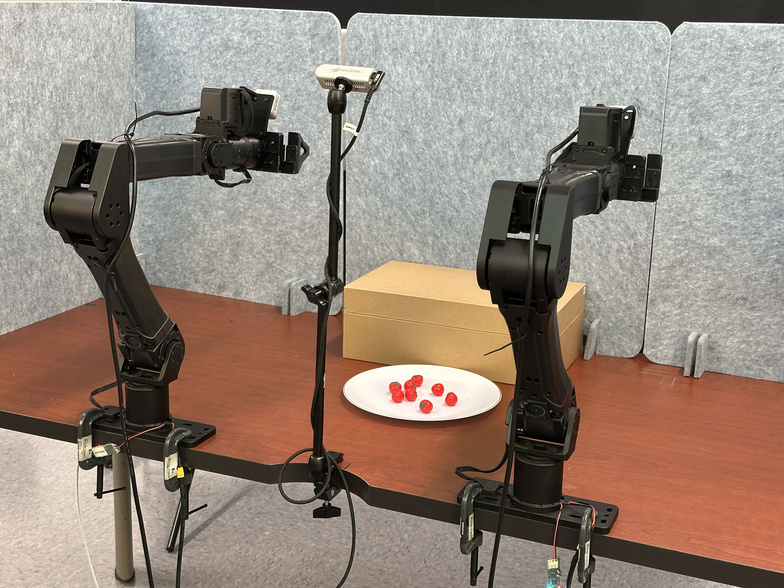}
    \end{minipage}
    \caption{\texttt{Tomato Plate\\Transfer}.}
    \label{fig:lift_plate}
\end{subfigure}\hfill
\begin{subfigure}[t]{0.194\linewidth}
    \centering
    \begin{minipage}[c][\yamPanelHeight][c]{\linewidth}
    \includegraphics[width=\linewidth,trim=120bp 0bp 434.4bp 0bp,clip]{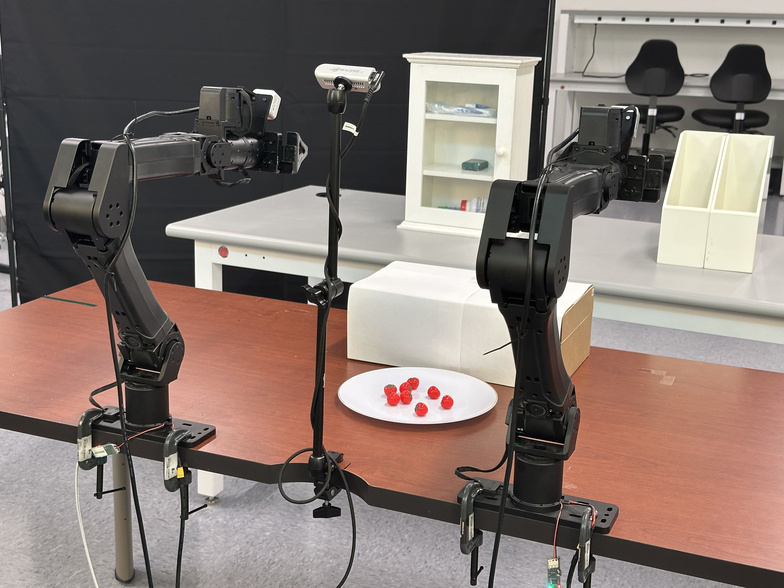}
    \end{minipage}
    \caption{\texttt{Tomato Plate\\Transfer (OOD)}.}
    \label{fig:lift_plate_ood}
\end{subfigure}\hfill
\begin{subfigure}[t]{0.40\linewidth}
    \centering
    \begin{minipage}[c][\yamPanelHeight][c]{\linewidth}
    \includegraphics[width=\linewidth]{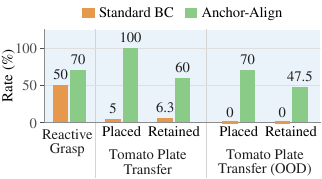}
    \end{minipage}
    \captionsetup{font=small}
    \caption{\baseline versus \ourmethod on \vlaadapter.}
    \label{fig:yam_results}
\end{subfigure}%
}
\caption{\small \textbf{Bimanual Coordination on YAM.}
\textbf{(a)} \texttt{Reactive Grasp}: a person carries a bottle across the robot's workspace, requiring the robot to grasp it while in motion.
\textbf{(b)} \texttt{Tomato Plate Transfer}: both arms lift a plate carrying eight tomatoes and place it onto a box without letting any tomatoes fall off.
\textbf{(c)} \texttt{Tomato Plate Transfer (OOD)}: the same task in an unseen environment, with the box covered in white paper and the surrounding workspace walls removed.
\textbf{(d)} Reactive Grasp success rate, plate-placement success rate (\emph{Placed}), and mean percentage of tomatoes retained (\emph{Retained}).}
\label{fig:control_tasks}
\end{figure}

\textbf{Bimanual Coordination.} In Anchor-Align, the \textit{Language-Action Alignment} loss is applied independently to each arm, with a separate motion-direction label for each. We compare Anchor-Align against \baseline across three tasks. 1) \texttt{Reactive Grasp} requires the robot to intercept and grasp a bottle carried by a moving person, adapting its reach and gripper closure to the bottle's motion. The person moves either from left to right or right to left, requiring the robot to switch its grasping arm based on the approach direction. Specifically, the left arm grasps the bottle when the person moves from right to left, and vice versa. The bottle varies in speed, height, and holding pose \citep{feng2026tempo}, requiring the policy to anticipate its motion rather than react only to its current position. 2) \texttt{Tomato Plate Transfer} requires the robot to lift a plate carrying eight tomatoes and place it onto a box without letting any tomatoes fall off. The robot must coordinate both arms to keep the plate level and maintain smooth motion, as differences in arm height can tilt the plate, while abrupt movements can cause the tomatoes to roll off. This task is challenging even for human teleoperators because both constraints must be maintained throughout the transfer \citep{song2026learning}. 3) \texttt{Tomato Plate Transfer (OOD)} evaluates the same task in an unseen environment, where the box is covered with white paper and the surrounding workspace walls are removed.

With VLA-Adapter on YAM, \ours improves \texttt{Reactive Grasp} success from 50\% to 70\%. On \texttt{Tomato Plate Transfer}, plate-placement success increases from 5\% to 100\%, while mean tomato retention increases from 6.3\% to 60\%. Under the unseen visual configuration, placement success increases from 0\% to 70\% and mean tomato retention from 0\% to 47.5\% (\cref{fig:yam_results}). These gains indicate substantially improved motion-sensitive control, spanning both reactive grasping of moving targets and coordinated bimanual transport. To better understand the bimanual coordination gains, we analyze inter-arm motion in Appendix~\ref{subsec:bimanual_height_coordination}, showing that \ourmethod reduces the mean inter-arm height-change mismatch by 46.2\%.

% \newpage

% \Needspace{20\baselineskip}
% \subsection{Analysis}
% \label{sec:analysis}
% \vspace{-0.15cm}

% % In this subsection, we empirically showcase 1) Anchor-Align preserves the VLM's pretrained representations; 2) existing co-trained VLAs exhibit language-action misalignment, whereas \ours improves alignment, which is positively correlated with action success.

% This section presents an in-depth analysis of language preservation and language-action misalignment in VLA training.

% \section{Conclusion}
% \label{sec:conclusion}
\section{Limitations and Error Analysis}
\label{sec:limitations}

\begin{wrapfigure}{r}{0.46\linewidth}
    \vspace{-\baselineskip}
    \centering
    \includegraphics[width=\linewidth]{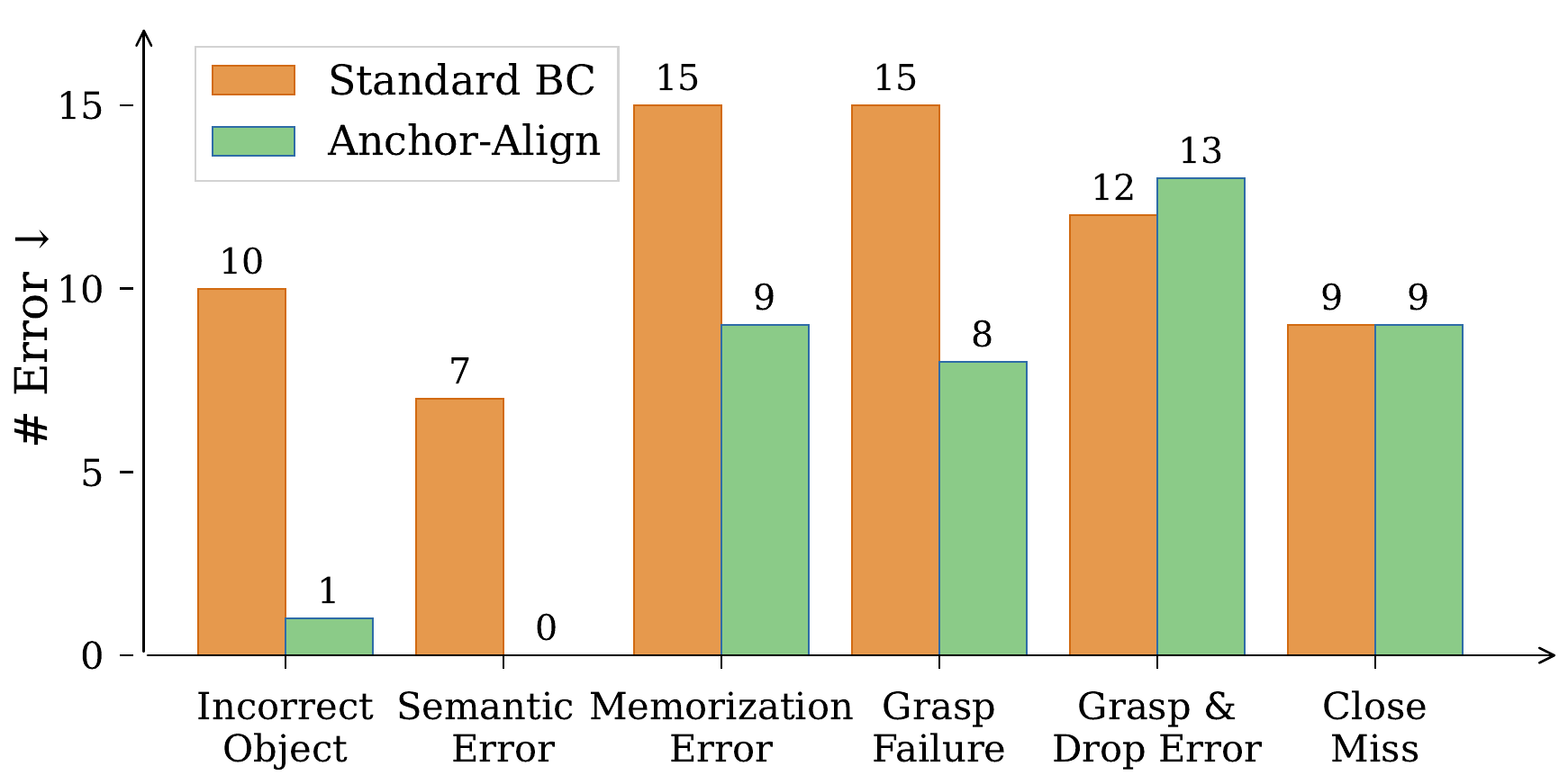}
    \caption{\small Failure mode analysis comparing standard BC and \ourmethod.}
    \label{fig:failure_modes}
    \vspace{-4pt}
\end{wrapfigure}

We manually inspect all failed xArm7 rollouts and bin clearly attributable failures into six modes (\cref{fig:failure_modes}): \emph{Incorrect Object}, approaches the wrong object; \emph{Semantic Error}, selects a feature-similar distractor (e.g., green grapes instead of green broccoli); \emph{Memorization Error}, ignores the scene and follows the training trajectory; \emph{Grasp Failure}, reaches the target but does not descend to grasp; \emph{Grasp-and-Drop Error}, grasps correctly but releases too early, too high, or not at all; and \emph{Close Miss}, grounds correctly but misses by a small margin. \cref{fig:failure_modes} shows that \ours eliminates semantic errors ($7\!\to\!0$), nearly eliminates incorrect-object approaches ($10\!\to\!1$), reduces memorization errors ($15\!\to\!9$), and nearly halves grasp failures ($15\!\to\!8$). The slight grasp-and-drop increase ($12\!\to\!13$) reflects more successful grasps, hence more opportunities for drop errors.

% \vspace{-10pt}
\section{Conclusion}
% \vspace{-5pt}
% \vspace{-5pt}

\label{sec:conclusion}
\noindent Standard BC can overwrite pretrained VLM semantics, while co-training does not ensure language-action alignment. \ourmethod addresses both through representation anchoring and language-action alignment losses, consistently improving OOD generalization, robustness, and real-world manipulation across VLA architectures and action heads without additional data or annotation. These results show that preserving and grounding VLM priors improves generalization and action learning.

% \noindent Finally, the principles of anchoring and alignment may extend beyond VLAs to video world models. For action-conditioned world models adapted from pretrained video generators~\citep{rigter2025avid,routray2025vipra,li2025uva,jang2025dreamgen}, frozen-teacher regularization could limit representation drift during robot-specific post-training. An analogous alignment objective could enforce action-outcome consistency (e.g., requiring an inverse-dynamics model to recover the conditioning action from the generated transition) and instruction-rollout consistency when language is present. This offers a concrete path toward world models that retain broad spatiotemporal priors while remaining controllable and semantically grounded.

%===============================================================================

\section*{Acknowledgments}
We thank R. Yeeshu Dhurandhar, Tongmiao Xu, Sagar Patil, and Zhenyang Feng for fruitful discussions. This research is based upon work supported by NSF grant CCF-2348624. This work also received support from the School of ICS at UC Irvine and from Modal through cloud compute credits and infrastructure. This research also used the DeltaAI advanced computing and data resource, which is supported by the National Science Foundation under award OAC 2320345 and the State of Illinois. DeltaAI is a joint effort of the University of Illinois Urbana-Champaign and its National Center for Supercomputing Applications. This work was also supported by the U.S. DARPA ECOLE Program under Award No.~HR00112390060. The views and conclusions contained herein are those of the authors and should not be interpreted as necessarily representing the official policies, either expressed or implied, of DARPA or the U.S. Government. The U.S. Government is authorized to reproduce and distribute reprints for governmental purposes, notwithstanding any copyright annotation therein.

%===============================================================================

% One reference list for the main paper and the appendix, placed before the appendix.
\bibliographystyle{iclr2027_conference}
\bibliography{main}

%===============================================================================
% Supplementary Material
\clearpage
\appendix
\raggedbottom
\hypertarget{anchor-align-appendix-start}{}
% \begin{center}
% {\Large \textbf{Training VLAs with Vision-Language Anchoring and Language-Action Alignment}}\\[1em]
{\large \textbf{Appendix}}
% \end{center}
% \vspace{0.5em}

% \noindent\textit{We submitted this supplementary material together with the main paper for ease of reference during review; here, we provide a slightly revised version for the supplementary deadline.}\\[0.4em]
% \noindent Project page: \url{https://anchorvla2026.github.io/AnchorVla2026/}
% \vspace{1em}

\noindent \ref{app:baseline_details}: \hyperref[app:baseline_details]{Baseline Details} \dotfill \pageref{app:baseline_details} \\
~\ref{app:implementation_detail}: \hyperref[app:implementation_detail]{Implementation Details} \dotfill \pageref{app:implementation_detail} \\
\hspace*{1.5em}\ref{subsec:align_head}~Language-Action Alignment Head \\
\hspace*{1.5em}\ref{subsec:hw_setup}~Training Setup \\
\hspace*{1.5em}\ref{subsec:train_cost}~Training Cost Comparison \\
\hspace*{1.5em}\ref{subsec:realworld_impl}~Real-World Implementation \\
\hspace*{1.5em}\ref{app:alignment_vs_reg}~Alignment vs.\ Regularization: Shuffle and Scatter Controls \\
~\ref{app:extra_results}: \hyperref[app:extra_results]{Extended Quantitative Results} \dotfill \pageref{app:extra_results} \\
\hspace*{1.5em}\ref{app:benchmark_subcategories}~LIBERO-PRO and LIBERO-Plus Subcategories \\
\hspace*{1.5em}\ref{subsec:standard_libero}~Success Rates on Standard LIBERO Suites \\
\hspace*{1.5em}\ref{subsec:per_suite_robustness}~Per-Suite Robustness Breakdowns \\
\hspace*{1.5em}\ref{app:multi_seed}~Multi-Seed Evaluation: Statistical Significance \\
\hspace*{1.5em}\ref{app:backbone_regularization}~Backbone Regularization Controls \\
~\ref{equations}: \hyperref[equations]{Representation Analysis: Language Preservation and Action Decodability} \dotfill \pageref{equations} \\
\hspace*{1.5em}\ref{subsec:cka}~Centered Kernel Alignment (CKA) \\
\hspace*{1.5em}\ref{subsec:action_decodability}~Action Decodability (Linear Probing $R^2$) \\
\hspace*{1.5em}\ref{subsec:preservation_decodability}~Language Preservation and Action Decodability \\
\hspace*{1.5em}\ref{subsec:layerwise_cka}~Layer-wise Language Preservation \\
\hspace*{1.5em}\ref{app:direction_understanding}~Direction Understanding in the Backbone \\
~\ref{sec:diagnostic}: \hyperref[sec:diagnostic]{Language-Action Alignment Diagnostic} \dotfill \pageref{sec:diagnostic} \\
\hspace*{1.5em}\ref{benchmark_curation}~Diagnostic Protocol \\
\hspace*{1.5em}\ref{reliability_of_language}~Relationship to Task Success \\
~\ref{real_world_rollouts}: \hyperref[real_world_rollouts]{Real-World Rollouts} \dotfill \pageref{real_world_rollouts} \\
\hspace*{1.5em}\ref{subsec:bimanual_height_coordination}~Bimanual Height Coordination \\
\hspace*{1.5em}\ref{subsec:object_orientation}~Object-Orientation Perturbation \\
\hspace*{1.5em}\ref{subsec:anchor_align_rollouts}~Anchor-Align Real-World Rollouts \\
~\ref{app:libero_rollouts}: \hyperref[app:libero_rollouts]{Qualitative Results in LIBERO Simulator} \dotfill \pageref{app:libero_rollouts}

\section{Baseline Details}
\label{app:baseline_details}

This appendix extends the condensed baseline list in \secref{sec:benchmarks}: baselines already introduced there are referenced briefly, and we detail the additional baselines that appear only in the appendix tables.

\noindent $\bullet$ \textbf{Standard VLM-based VLAs.}
OpenVLA-OFT~\citep{kim2025openvlaoft} and \vlaadapter~\citep{wang2025vla} (our base model) are introduced in \secref{sec:benchmarks}. The standard-LIBERO comparison (\cref{tab:standard_success}) additionally includes VLA-0~\citep{goyal2025vla0} (plain-text action integers).

\noindent $\bullet$ \textbf{Knowledge-preservation approaches.}
$\pi_{0.5}$-KI~\citep{driess2025knowledge} addresses catastrophic forgetting by training discrete FAST action tokens on the VLM backbone while stop-gradient prevents action gradients from corrupting pretrained representations. \frozen~\citep{wang2025vla} represents the extreme preservation baseline: the VLM backbone is entirely frozen and only the action head is trained, trivially preserving all pretrained knowledge at the cost of reduced task performance. MAPS~\citep{huang2025maps} instead keeps each module's weights close to the pretrained checkpoint through a layer-wise proximity schedule that constrains vision layers more tightly than action-oriented ones. Our MAPS baseline runs the authors' released proximity optimizer and schedule on \vlaadapter with the maximum proximity weight of $0.5$ that the MAPS authors use for \vlaadapter on LIBERO; all other settings follow the standard BC recipe in \appref{subsec:hw_setup}, and we evaluate the final 10K-step checkpoint.

\noindent $\bullet$ \textbf{Co-training baselines.}
We compare \textit{MolmoAct} and \textit{Co-training + KI}, introduced in \secref{sec:benchmarks}. MolmoAct~\citep{lee2025molmoact} is an example of co-training with general-purpose web examples separate from robot demonstrations. Its robot-side supervision jointly models depth-perception tokens, image-space trajectory traces, and discretized actions from the same observations for 3D spatial reasoning. \textit{Co-training + KI} is our reimplementation of Knowledge Insulation~\citep{driess2025knowledge} on \vlaadapter. We sample web VQA and robot action data at a ratio of $1{:}3$ (web:action). Action finetuning uses LoRA: the language modeling head receives VQA supervision on web examples and FAST-tokenized action supervision on robot observations, while an L1 regression action head predicts continuous actions. Knowledge insulation blocks gradients from the continuous action head into the backbone. These comparisons test whether co-training, including action-token supervision and knowledge insulation, is sufficient to preserve pretrained semantics and produce consistent language and action predictions.

\emph{Co-training during task-specific finetuning.} Co-training has shown strong results in large-scale VLA pretraining~\citep{lee2025molmoact}, and Knowledge Insulation improves knowledge transfer during VLA training~\citep{driess2025knowledge}. Our VLM-to-VLA finetuning setting emphasizes learning fine-grained task-specific behavior. We hypothesize that general-purpose web supervision can act as an auxiliary objective with limited relevance to these task details, potentially competing with task-specific adaptation. This may help explain why our Co-training + KI baseline underperforms standard BC on mean LIBERO-PRO and LIBERO-Plus success (\cref{tab:libero_suite}).

\noindent $\bullet$ \textbf{Long-horizon CALVIN baselines.}
The main-paper CALVIN ABC$\rightarrow$D comparison (\cref{tab:calvin_abcd}) includes OpenVLA-OFT~\citep{kim2025openvlaoft}, UniVLA~\citep{bu2025univla}, OpenHelix~\citep{cui2025openhelix}, and \vlaadapter~\citep{wang2025vla}. These baselines evaluate long-horizon, instruction-conditioned manipulation under environment shift.

\section{Implementation Details}
\label{app:implementation_detail}

For better reproducibility of our results, we include the full architectural specification, hyperparameter settings, and training cost breakdown below.

\subsection{Language-Action Alignment Head}
\label{subsec:align_head}
The alignment objective (Sec.~\ref{sec:lang-action-align}) introduces a single learned linear projection $\mathbf{W}_{\text{proj}} \in \mathbb{R}^{d \times d}$ with $d = 896$ (the hidden dimension of Qwen2.5-0.5B), adding $896^2 + 896 = 803{,}712$ parameters ($1.5$\,MB in \texttt{bfloat16}). At each training step, we take the last-layer hidden state $\mathbf{h}_i^{\text{pre}} = \mathbf{H}^{S}[N] \in \mathbb{R}^d$ of the last of the $N$ vision and text tokens, the position immediately before the first action token (located in code by scanning the label tensor for the first non-masked entry, which marks the start of the action tokens). It is then mapped through $\mathbf{W}_{\text{proj}}$ and the frozen pretrained language modeling head $\mathbf{W}_{\text{lm}} \in \mathbb{R}^{|\mathcal{V}| \times d}$ to produce the language logits $\hat{\mathbf{a}}^{\text{lang}}_i \in \mathbb{R}^{|\mathcal{V}|}$ over the full vocabulary ($|\mathcal{V}| = 151{,}646$). A cross-entropy loss is then computed between $\hat{\mathbf{a}}^{\text{lang}}_i$ and the ground-truth direction label $\mathbf{a}^{\text{lang}}_i$ (one of \emph{forward}, \emph{backward}, \emph{left}, \emph{right}, \emph{up}, \emph{down}), with near-stationary samples ($\|\bar{\mathbf{v}}_i\|_2 < \tau$, $\tau = 0.15$) masked via \texttt{ignore index}. Because $W_{\mathrm{lm}}$ is frozen, the alignment gradient flows only through $W_{\mathrm{proj}}$ and the backbone's LoRA parameters; the pretrained vocabulary head therefore remains fixed throughout
alignment training. The projection head is the only additional trainable module beyond the LoRA adapters and action head already present in \vlaadapter.

\subsection{Training Setup}
\label{subsec:hw_setup}
All models are trained on 4 NVIDIA GH200 GPUs using \texttt{bfloat16} mixed precision with PyTorch~2.7.

\noindent\textbf{Hyperparameters.}
To ensure reproducibility, we report the complete training configuration used for the results in \cref{tab:libero_suite}.
The backbone VLM receives two camera images (third-person and wrist) along with the proprioceptive state as input.
We use LoRA-based parameter-efficient finetuning with rank $r{=}64$ and scaling factor $\alpha{=}128$, applied to all linear layers in the backbone, and merge the LoRA weights into the base model periodically during training.
Training runs for 10{,}000 gradient steps with a batch size of 32, a learning rate of $2 \times 10^{-4}$, and standard image augmentation (random crops and color jitter).
Checkpoints are saved every 2{,}500 steps.
For the anchoring objective, we apply layer-wise MSE distillation ($\Lanchor$) across all 24 transformer layers with anchoring weight $\lambda_{\text{anchor}}{=}0.05$.
For the alignment objective ($\Lalign$), we use an alignment weight of $\lambda_{\text{align}}{=}0.02$ and a motion-magnitude threshold $\tau{=}0.15$ to filter near-stationary samples.
All remaining hyperparameters, including optimizer settings and data preprocessing, follow the recommended defaults from ~\citep{wang2025vla} for the baseline setup.

\noindent\textbf{Image encoding.}
The $512$ vision patches are obtained from two images: we extract DINOv2~\citep{oquab2023dinov2} and SigLIP~\citep{zhai2023sigmoid} features for each image and concatenate them along the feature dimension per patch, yielding 256 patches per image and a 512-patch vision prefix in total.

\subsection{Training Cost Comparison}
\label{subsec:train_cost}
A key practical question is how much overhead knowledge-preservation methods add to standard action finetuning. On identical hardware, we compare \ourmethod against \textbf{Standard BC} (\vlaadapter~\citep{wang2025vla}) and \textbf{Co-training + KI}~\citep{driess2025knowledge}, both detailed in \secref{sec:benchmarks}. Co-training is the natural point of comparison because it explicitly targets the same goal as our method: retaining the VLM's pretrained visuolinguistic priors during finetuning. \cref{tab:train_cost} summarizes the cost and performance trade-offs.

\begin{table}[H]
    \centering
    \small
    \setlength{\tabcolsep}{4pt}
    \renewcommand{\arraystretch}{1.15}
    \caption{Training cost and downstream performance comparison. The recorded profiling measurements show $3.4\times$ less overhead per step and $7\times$ less additional GPU memory for \ourmethod, with no external data. KI timing and memory measurements are from the earlier profiling run; its success rate is taken from \cref{tab:libero_suite}.}
    \label{tab:train_cost}
\begin{tabular}{@{}l c c c c@{}}
    \toprule
    & \textbf{Standard BC} & \textbf{Co-training + KI\footnotemark} & \textbf{\ourmethod} \\
    \midrule
    Wall-clock (s/it)         & 1.28  & 2.50  & 1.64  \\
    Overhead vs.\ baseline    & ---   & +95\% & +28\% \\
    2,500 steps (min)         & 53    & 104   & 68    \\
    Extra GPU memory          & ---   & +5\,GB & +0.7\,GB \\
    External data required    & None  & 25K VQA & None  \\
    \midrule
    LIBERO-Plus Overall       & 85.1  & 72.4  & \textbf{90.3}  \\
    \bottomrule
    \end{tabular}
    
\end{table}
\footnotetext{Our reimplementation on \vlaadapter (\secref{sec:benchmarks}).}

\noindent
In the earlier profiling run, Co-training + KI adds a full VQA forward and backward pass (with batch size 16) at every training step, incurring $+95\%$ wall-clock overhead and $+5$\,GB of GPU memory for the additional activations and gradients. It further requires curating 25K external VQA samples (VSR \citep{liu-etal-2023-visual}, GQA \citep{hudson2019gqa}, COCO \citep{lin2014microsoft}) and maintaining the language modeling head alongside the action head during training. Co-training + KI scores 53.6\% on LIBERO-PRO language rephrase and 72.4\% LIBERO-Plus overall, below standard BC (91.1\% and 85.1\%) in these evaluations.

In contrast, \ourmethod adds only a single inference-only forward pass through a frozen copy of the pretrained VLM (the \emph{anchor}). We replace the anchor's language modeling head ($151{,}646 \times 896 \approx 136$M parameters) with \texttt{nn.Identity()}, since only intermediate hidden states are needed for the layer-wise MSE loss, reducing the anchor's footprint to $0.7$\,GB. This yields $+28\%$ wall-clock overhead ($0.36$\,s per step), requires no external data or annotation, and completes a full 2{,}500-step run in $68$ minutes---$36$ minutes faster than co-training. Critically, \ourmethod achieves 97.0\% on language rephrase and 90.3\% on LIBERO-Plus overall, demonstrating that preserving the VLM's representational geometry via anchoring is both cheaper and dramatically more effective than co-training on external VQA data. For larger backbone VLMs, the relative overhead would decrease further, since the anchor's inference-only forward pass scales sublinearly compared to the training backward pass.

\subsection{Real-World Implementation}
\label{subsec:realworld_impl}

\noindent Throughout our real-world experiments, ``standard BC'' denotes the published \vlaadapter~\citep{wang2025vla} and StarVLA~\citep{starvla2026} architectures trained with their respective behavior-cloning objectives, using the same demonstrations and optimization settings as \ours within each architecture.

\noindent\textbf{xArm7 Setup.}
We uses xArm7 UFactory equipped with a wrist-mounted RGB camera and an external front-view camera, evaluated on tabletop tasks requiring object discrimination, spatial reasoning, and instruction-conditioned target selection under clutter. We collect 50 demonstration trajectories per task using a 3D-printed GELLO teleoperation device~\citep{wu2024gello} with velocity control at 30\,Hz; trajectories are stored in the LeRobot~\citep{cadene2024lerobot} format and converted to RLDS~\citep{ramos2021rlds} for our training pipeline. Both baseline and \ours are trained with identical hyperparameters: \vlaadapter policies for 10K steps and StarVLA policies for 50K steps. Evaluation follows the task definitions and rollout protocol in Sec.~\ref{sec:real_world}. At deployment, the low-level velocity controller runs at 100\,Hz while the learned policy predicts actions at 30\,Hz; commands are interpolated and velocity-clipped for safe execution.

\noindent\textbf{Bimanual Robot Setup.}
We conduct our bimanual experiments on an I2RT YAM Ultra platform comprising two 6-DoF arms, each equipped with a 1-DoF parallel-jaw gripper. We collect demonstrations through VR teleoperation using a Meta Quest Pro headset and handheld controllers. The controllers command arm motion and gripper opening and closing.

\noindent\textbf{Bimanual Training.}
We train on 90 demonstrations for \texttt{Reactive Grasp} and 19 successful demonstrations for \texttt{Tomato Plate Transfer}; each task is trained separately on all of its demonstrations with no held-out episodes, and \baseline and \ourmethod are trained on the same demonstrations. Both are trained with \vlaadapter for 10K steps with identical hyperparameters.

\noindent\textbf{Bimanual direction labels.}
YAM has two arms, so its 14-dimensional action concatenates two 7-DoF arm actions, each with a translational component $(\Delta x, \Delta y, \Delta z)$, a rotational component, and a gripper command. The two arm bases are parallel, so we simply duplicate the label construction of Sec.~\ref{sec:lang-action-align}: the dominant-axis rule is applied to each arm's $(\Delta x, \Delta y, \Delta z)$ separately, with the same axis-to-word mapping for both arms. Each arm thus receives its own direction word (e.g., \emph{up} for both arms while lifting the plate), and the alignment loss is averaged over the two arms.

\subsection{Alignment vs.\ Regularization: Shuffle and Scatter Controls}
\label{app:alignment_vs_reg}

\noindent This section gives the full construction of the \textit{Shuffle} and \textit{Scatter} controls; their motivation and headline results are covered in \secref{sec:ablation} and \cref{tab:shuffle_control_app}. Both controls are trained with hyperparameters identical to \ours: the only quantity that changes is the lookup table from motion class to target token id, so here we are destroying semantic content in these controls.

\noindent\textbf{\textit{Shuffle}.} The observation-to-label mapping is reshuffled once and held fixed throughout training. Concretely, we apply a fixed derangement of the six direction words: \textit{forward}\,$\to$\,\texttt{right}, \textit{backward}\,$\to$\,\texttt{down}, \textit{left}\,$\to$\,\texttt{forward}, \textit{right}\,$\to$\,\texttt{up}, \textit{up}\,$\to$\,\texttt{left}, and \textit{down}\,$\to$\,\texttt{backward}.

\noindent\textbf{\textit{Scatter}.} Each direction class is mapped one-to-one to a fixed meaningless word: \textit{forward}\,$\to$\,\texttt{purple}, \textit{backward}\,$\to$\,\texttt{seven}, \textit{left}\,$\to$\,\texttt{today}, \textit{right}\,$\to$\,\texttt{music}, \textit{up}\,$\to$\,\texttt{tree}, and \textit{down}\,$\to$\,\texttt{math}. The six words are chosen to be mutually distant in the frozen output-embedding space, and the targets remain perfectly synchronized with the observations, so the model still solves a well-defined six-way classification task; only the motion semantics of the target tokens are removed.

\noindent Per axis, the collapse reported in \cref{tab:shuffle_control_app} is sharpest on the LIBERO-PRO position-swap axis, where success requires re-grounding language to the perturbed scene rather than memorizing scene-to-action mappings: \ours achieves $22.6\%$ on position swap while \textit{Shuffle} and \textit{Scatter} recover only the baseline's near-zero performance.

\section{Extended Quantitative Results}
\label{app:extra_results}

This section complements the main paper's quantitative evaluation (Sec.~\ref{sec:quantitative}) with benchmark subcategory details, success rates on the standard LIBERO suites, per-suite robustness breakdowns, multi-seed significance tests, and backbone regularization controls.
\label{subsec:extended_suites}

\subsection{LIBERO-PRO and LIBERO-Plus Subcategory Clarifications}
\label{app:benchmark_subcategories}

LIBERO-PRO language rephrase rewords the instruction (``pick up the mug'' becomes ``grab the mug''), whereas LIBERO-Plus language instruction rewrites it globally, e.g., referring to the target by its function (``the flat surface used for holding food'' for a plate), forcing the policy to ground the words in the observed scene. Similarly, LIBERO-PRO object swap replaces scene objects with unseen variants and position swap exchanges their locations (\cref{fig:swap_qualitative}), whereas LIBERO-Plus object layout keeps the objects but shifts the target location and adds distractors.

\subsection{Success Rates on Standard LIBERO Suites}
\label{subsec:standard_libero}

We report success rates on the four standard (unperturbed) LIBERO suites---Spatial, Object, Goal, and Long---in \cref{tab:standard_success}. \ours achieves the highest success rate on most suites, surpassing prior methods including $\pi_{0.5}$-KI and OpenVLA-OFT, which use substantially larger backbones and large-scale robotic pretraining. This confirms that the gains from anchoring and alignment hold not only on the OOD stress tests reported in the main paper (\cref{tab:libero_suite}), but also on the in-distribution benchmark.

\begin{table}[t]
\centering
\scriptsize
\setlength{\tabcolsep}{2.5pt}
\renewcommand{\arraystretch}{1.1}
\caption{\textbf{Success rates across standard LIBERO suites.} Anchoring and alignment achieve state-of-the-art results on unperturbed (standard) benchmarks. Best per column is \textbf{bolded}.}
\label{tab:standard_success}
\begin{adjustbox}{max width=\linewidth}
\begin{tabular}{@{}l cccc@{}}
    \toprule
    \textbf{Method} &
    \cellcolor{LightBlue}\textbf{Spatial} & \cellcolor{LightRed}\textbf{Object} & \cellcolor{LightYellow}\textbf{Goal} & \cellcolor{LightPurple}\textbf{Long} \\
    \midrule
    OpenVLA-OFT~\citep{kim2025openvlaoft} & 94.3 & 95.2 & 91.7 & 86.5 \\
    \molmoact~\citep{lee2025molmoact} & 87.0 & 95.4 & 87.6 & 77.2 \\
    $\pi_{0.5}$-KI~\citep{driess2025knowledge} & 96.6 & 97.2 & 94.6 & 85.8 \\
    VLA-0~\citep{goyal2025vla0} & 93.6  & 96.0 & 95.6 & 87.6 \\
    \frozen~\citep{wang2025vla}   & 89.4  & 89.6    & 88.0    & 84.5   \\
    \vlaadapter~\citep{wang2025vla}       & 96.0  & \textbf{99.8}  & 96.0  & 89.0 \\
    \rowcolor{LightGreen}
    \ours      & \textbf{98.4} & \textbf{99.8} & \textbf{97.2} & \textbf{90.8} \\
    \bottomrule
\end{tabular}%
\end{adjustbox}
\end{table}

\subsection{Per-Suite Robustness Breakdowns}
\label{subsec:per_suite_robustness}

\begin{figure}[t]
    \centering
    \includegraphics[width=0.65\linewidth]{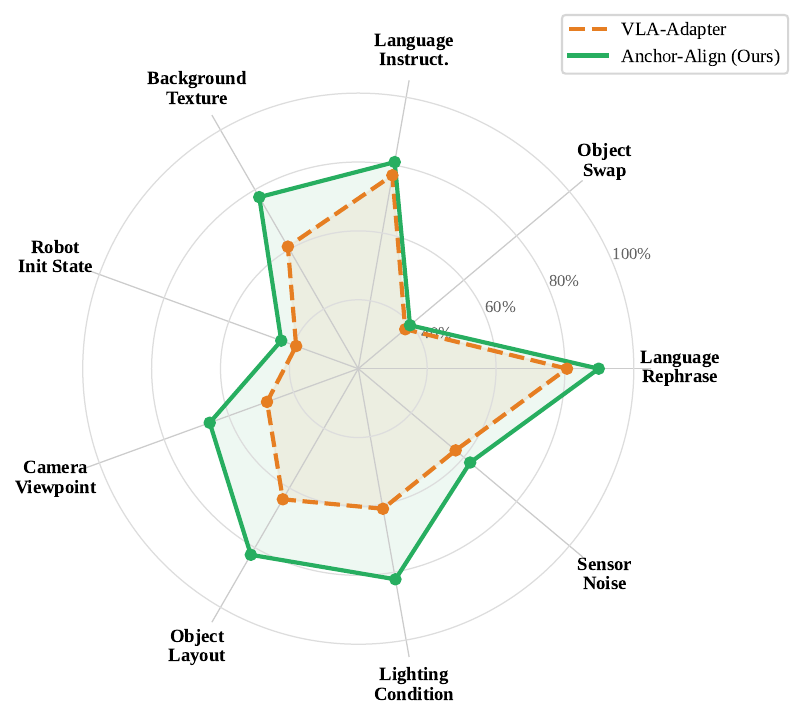}
    \caption{\small \textbf{LIBERO Long Suite.} Performance across nine evaluation axes (excluding Standard and Position Swap). \ours (green) substantially expands coverage over \vlaadapter (orange), with the largest gains on Lighting Condition (+20.8\%), Object Layout (+18.6\%), and Camera Viewpoint (+17.7\%).}
    \label{fig:long_spider}
    \vspace{-10pt}
\end{figure}

\begin{figure}[t]
    \centering
    \includegraphics[width=0.55\linewidth]{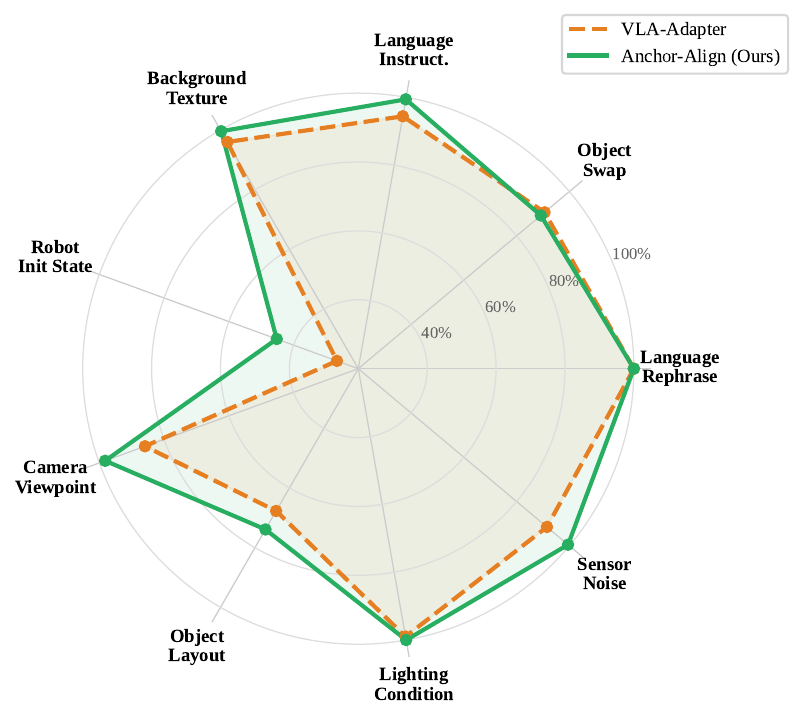}
    \caption{\small \textbf{LIBERO Object Suite.} Performance across nine evaluation axes (excluding Standard and Position Swap). \ours (green) consistently covers a larger area than \vlaadapter (orange), with the most pronounced gains on Robot Init State (+18.6\%).}
    \label{fig:object_spider}
    \vspace{4pt}
    \includegraphics[width=0.55\linewidth]{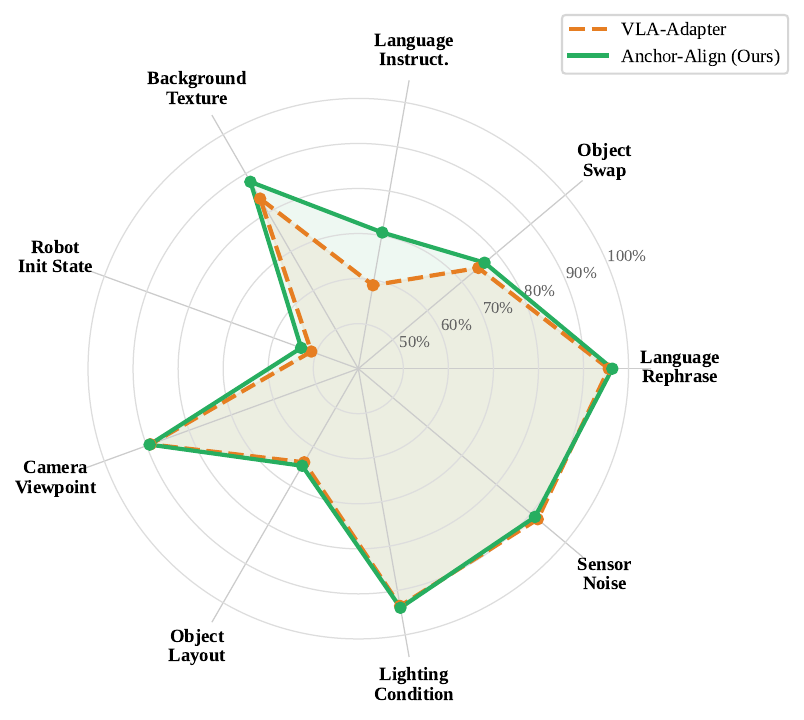}
    \caption{\small \textbf{LIBERO Goal Suite.} Performance across nine evaluation axes (excluding Standard and Position Swap). \ours (green) shows the largest gains over \vlaadapter (orange) on Language Instruction (+11.9\%) and Background Texture (+4.3\%).}
    \label{fig:goal_spider}
\end{figure}

In the main paper (\cref{tab:libero_suite}), we report robustness and generalization results on the LIBERO-Spatial suite, where perturbations to object positions, language instructions, and scene layout directly test whether the policy has learned transferable spatial reasoning or merely memorized training trajectories. Here we show that the same gains carry over to the remaining three LIBERO suites: Long (\cref{fig:long_spider}), Object (\cref{fig:object_spider}), and Goal (\cref{fig:goal_spider}), which test long-horizon composition, object knowledge, and procedural reasoning, respectively. All comparisons are against \vlaadapter, the strongest standard BC baseline from \cref{tab:libero_suite}. Because \vlaadapter uses the same LoRA architecture and training data as \ours but without anchoring or alignment, it directly isolates the effect of our two proposed objectives. Each radar plot reports performance across nine evaluation axes: two from LIBERO-PRO (Language Rephrase and Object Swap) and seven from LIBERO-Plus (Language Instruction, Background Texture, Robot Init State, Camera Viewpoint, Object Layout, Lighting Condition, and Sensor Noise). Standard and Position Swap axes are excluded.

\subsection{Multi-Seed Evaluation: Statistical Significance}
\label{app:multi_seed}

To confirm that the gains reported in the main paper's \cref{tab:libero_suite} (LIBERO-PRO and LIBERO-Plus) are not artifacts of a single random seed, we re-run both \ours and the \vlaadapter baseline with five independent training seeds and report mean $\pm$ standard deviation in \cref{tab:multi_seed}. The standard deviations are small (all $\leq 0.7$) and the mean improvements of \ours over \vlaadapter on every column are substantially larger than the combined seed-to-seed variability, so the differences are statistically significant by a paired comparison. The multi-seed means also closely match the single-seed numbers reported in \cref{tab:libero_suite}, indicating that the headline results are stable rather than seed-lucky.

\begin{table}[h]
\centering
\setlength{\tabcolsep}{8pt}
\renewcommand{\arraystretch}{1.15}
\footnotesize
\caption{\textbf{Multi-seed evaluation (5 seeds).} Mean $\pm$ standard deviation of success rate (\%) for \ours and the \vlaadapter baseline on LIBERO-PRO language rephrase / object swap / position swap, and the LIBERO-Plus overall average. Seed-to-seed variability is small ($\leq 0.7$) and substantially smaller than the gap between methods, confirming that the improvements reported in the main paper's \cref{tab:libero_suite} are statistically significant.}
\label{tab:multi_seed}
\begin{tabular}{l|ccc|c}
\toprule
\textbf{Method} & \textbf{Lang.\ Reph.} & \textbf{Object Swap} & \textbf{Pos.\ Swap} & \textbf{Plus} \\
\midrule
\vlaadapter & $91.1 \pm 0.4$ & $90.1 \pm 0.5$ & $2.6 \pm 0.7$ & $85.3 \pm 0.3$ \\
\ours & $\mathbf{97.1 \pm 0.5}$ & $\mathbf{96.1 \pm 0.4}$ & $\mathbf{23.5 \pm 0.2}$ & $\mathbf{90.5 \pm 0.6}$ \\
\bottomrule
\end{tabular}
\end{table}

\noindent\textbf{Gains transfer to a larger VLM backbone.}
To verify that the improvements from \ourmethod are not specific to our default backbone or action head, we re-instantiate the method on a different StarVLA configuration than the one used in the main-paper real-world experiments (\secref{sec:real_world}). Recall that StarVLA is a modular VLA framework that combines an arbitrary VLM backbone with an arbitrary action head; here we instantiate it with a larger Qwen2.5~3B VLM and an OFT MLP-based action head, in contrast to the Qwen2.5-VL + GR00T FM-DiT instantiation used in the main paper. On LIBERO-PRO, \ours improves the StarVLA baseline from $73.6\%$ to $\mathbf{89.0\%}$ on the language rephrase axis and from $89.8\%$ to $\mathbf{91.2\%}$ on the object swap axis. The improvement is qualitatively consistent with what we observe on the smaller backbone, indicating that the benefits of anchoring and alignment are a property of the training objective rather than of a particular VLM scale or action-head design.

\subsection{Backbone Regularization Controls}
\label{app:backbone_regularization}

We compare Vision-Language Anchoring against five alternative backbone regularization settings: reducing the LoRA rank to $r=32$, reducing the learning rate by a factor of ten, post-hoc interpolation between the adapted and frozen backbone weights at $\alpha=0.9$ and $\alpha=0.75$, and applying adapter weight decay of $0.1$. As shown in \cref{tab:regularization_comparison}, anchoring alone achieves the highest mean success rate on both LIBERO-PRO and LIBERO-Plus, exceeding the strongest alternative, adapter weight decay, by $5.7$ and $1.7$ percentage points, respectively. These results support explicitly preserving pretrained representations through anchoring over the tested controls that limit adaptation through capacity, optimization, or weight-space regularization.

\begin{table}[htbp]
\centering
\footnotesize
\setlength{\tabcolsep}{8pt}
\renewcommand{\arraystretch}{1.1}
\caption{\textbf{Backbone regularization comparison.} Mean success rates (\%) on LIBERO-PRO (PRO) and LIBERO-Plus (Plus). Adapter$\leftrightarrow$frozen denotes post-hoc adapter scaling, equivalent to weight-space interpolation with the frozen backbone. Anchoring alone outperforms all five alternative settings on both benchmarks.}
\label{tab:regularization_comparison}
\begin{tabular}{@{}lcc@{}}
\toprule
\textbf{Backbone regularization} & \textbf{PRO} & \textbf{Plus} \\
\midrule
Reduced LoRA rank ($r=32$) & 60.0 & 80.4 \\
Reduced learning rate ($\times 0.1$) & 55.8 & 71.3 \\
Adapter$\leftrightarrow$frozen ($\alpha=0.9$) & 61.8 & 85.0 \\
Adapter$\leftrightarrow$frozen ($\alpha=0.75$) & 59.0 & 82.5 \\
Adapter weight decay ($0.1$) & 62.4 & 85.6 \\
\midrule
\rowcolor{LightGreen}
\anchor (anchoring only) & \textbf{68.1} & \textbf{87.3} \\
\bottomrule
\end{tabular}
\end{table}

\section{Representation Analysis: Language Preservation and Action Decodability}
\label{equations}

This appendix provides the full mathematical details for the evaluation metrics used in the paper, followed by the representation-level analyses built on them.

\subsection{Centered Kernel Alignment (CKA)}
\label{subsec:cka}

We use CKA~\citep{kornblith2019similarity, bo2024evaluating} to measure how much of the pretrained VLM's representational geometry is preserved after finetuning.  CKA is built on the Hilbert--Schmidt Independence Criterion (HSIC)~\citep{gretton2005measuring}, which quantifies the statistical dependence between two sets of representations via their kernel matrices.

\noindent\textbf{HSIC.}
Given $n$ data points and two representation matrices $\mathbf{X} \in \mathbb{R}^{n \times p}$ and $\mathbf{Y} \in \mathbb{R}^{n \times q}$, we form the Gram matrices $\mathbf{K} = \mathbf{X}\mathbf{X}^\top$ and $\mathbf{L} = \mathbf{Y}\mathbf{Y}^\top$ (under a linear kernel).  The empirical HSIC is
\begin{equation}
    \mathrm{HSIC}(\mathbf{K}, \mathbf{L}) = \frac{1}{(n-1)^2}\,\mathrm{tr}\!\bigl(\mathbf{K}\,\mathbf{H}\,\mathbf{L}\,\mathbf{H}\bigr),
\end{equation}
where $\mathbf{H} = \mathbf{I}_n - \frac{1}{n}\mathbf{1}\mathbf{1}^\top$ is the centering matrix.  Centering removes the mean from each kernel matrix, so that HSIC captures the structural (second-order) similarity rather than mere magnitude.

\noindent\textbf{CKA.}
CKA normalizes HSIC so that the score lies in $[0,1]$:
\begin{equation}
    \mathrm{CKA}(\mathbf{K}, \mathbf{L}) = \frac{\mathrm{HSIC}(\mathbf{K}, \mathbf{L})}{\sqrt{\mathrm{HSIC}(\mathbf{K}, \mathbf{K})\;\mathrm{HSIC}(\mathbf{L}, \mathbf{L})}}.
\end{equation}
A CKA score of 1.0 indicates that the two representations have identical geometric structure up to an isotropic scaling and orthogonal transformation; lower values indicate that finetuning has reshaped the layer's representations.

\noindent\textbf{Linear CKA.}
When the linear kernel is used, the CKA expression simplifies to a closed-form that avoids explicitly constructing the $n \times n$ Gram matrices.  Let $\tilde{\mathbf{X}}$ and $\tilde{\mathbf{Y}}$ denote the column-centered versions of $\mathbf{X}$ and $\mathbf{Y}$.  Then:
\begin{equation}
\label{eq:linear_cka}
    \mathrm{CKA}_{\text{linear}}(\mathbf{X}, \mathbf{Y}) = \frac{\bigl\| \tilde{\mathbf{Y}}^\top \tilde{\mathbf{X}} \bigr\|_F^2}{\bigl\| \tilde{\mathbf{X}}^\top \tilde{\mathbf{X}} \bigr\|_F \;\cdot\; \bigl\| \tilde{\mathbf{Y}}^\top \tilde{\mathbf{Y}} \bigr\|_F}.
\end{equation}
This is the form we use throughout: at each transformer layer $u$, $\mathbf{X} \in \mathbb{R}^{n \times d}$ contains the text-token hidden states from the frozen pretrained VLM and $\mathbf{Y} \in \mathbb{R}^{n \times d}$ from the finetuned backbone VLM, across $n$ evaluation samples.  We report text-token CKA because it directly reflects the degree to which the model's language understanding is preserved.

\subsection{Action Decodability (Linear Probing $R^2$)}
\label{subsec:action_decodability}

To measure how much action-relevant information is linearly accessible from the model's hidden states, we train a ridge regression probe at each layer.  For layer $u$, we extract the hidden state and mean-pool over all token positions to obtain a feature vector $\mathbf{f}_i^{(u)} \in \mathbb{R}^{d}$ for each sample $i$.  We then fit a ridge regression model $g$ to predict the ground-truth discretized action $\mathbf{a}_i \in \mathbb{R}^{7}$ (with 256 bins per dimension):
\begin{equation}
    R^2(u) = 1 - \frac{\sum_{i=1}^{n} \bigl\| \mathbf{a}_i - g\bigl(\mathbf{f}_i^{(u)}\bigr) \bigr\|_2^2}{\sum_{i=1}^{n} \bigl\| \mathbf{a}_i - \bar{\mathbf{a}} \bigr\|_2^2},
\end{equation}
where $\bar{\mathbf{a}}$ is the mean action over the dataset.  A higher $R^2$ indicates that more action information is linearly decodable from that layer's representations, even without the action head.

\subsection{Language Preservation and Action Decodability}
\label{subsec:preservation_decodability}

We quantify what each training paradigm does to the backbone's internal representations along two axes: language preservation, measured with Centered Kernel Alignment (CKA)~\citep{bo2024evaluating} between the finetuned backbone's text-token hidden states and the pretrained VLM's (formal definition in App.~\ref{subsec:cka}), and action decodability, measured by linear-probing the backbone's hidden states for the ground-truth action (details in App.~\ref{subsec:action_decodability}). As shown in \cref{fig:preservation_decodability}, standard BC catastrophically destroys the pretrained text representations (CKA drops to 0.34; per-layer analysis in App.~\ref{subsec:layerwise_cka}). At the opposite extreme, the frozen backbone trivially preserves the pretrained geometry yet gains no action information, and the co-trained VLA behaves similarly. \ours achieves the best of both: it maintains strong language preservation (CKA = 0.91) while attaining the highest action decodability of any method (peak R$^2$ = 0.60 at layer 22), showing that alignment routes action-relevant information through the decoder layers without overwriting the pretrained language geometry. Preservation captures what the model keeps from pretraining, decodability what it gains from finetuning; together they show that \ours attains high task performance by enriching the pretrained representations rather than destroying them, as standard BC does. Additional linear-probe results on direction concepts (in-distribution vs.\ OOD) are reported in App.~\ref{app:direction_understanding}.

\subsection{Layer-wise Language Preservation}
\label{subsec:layerwise_cka}

\begin{figure}[H]
    \centering
    \begin{minipage}[t]{0.45\linewidth}
        \centering
        \includegraphics[width=\linewidth]{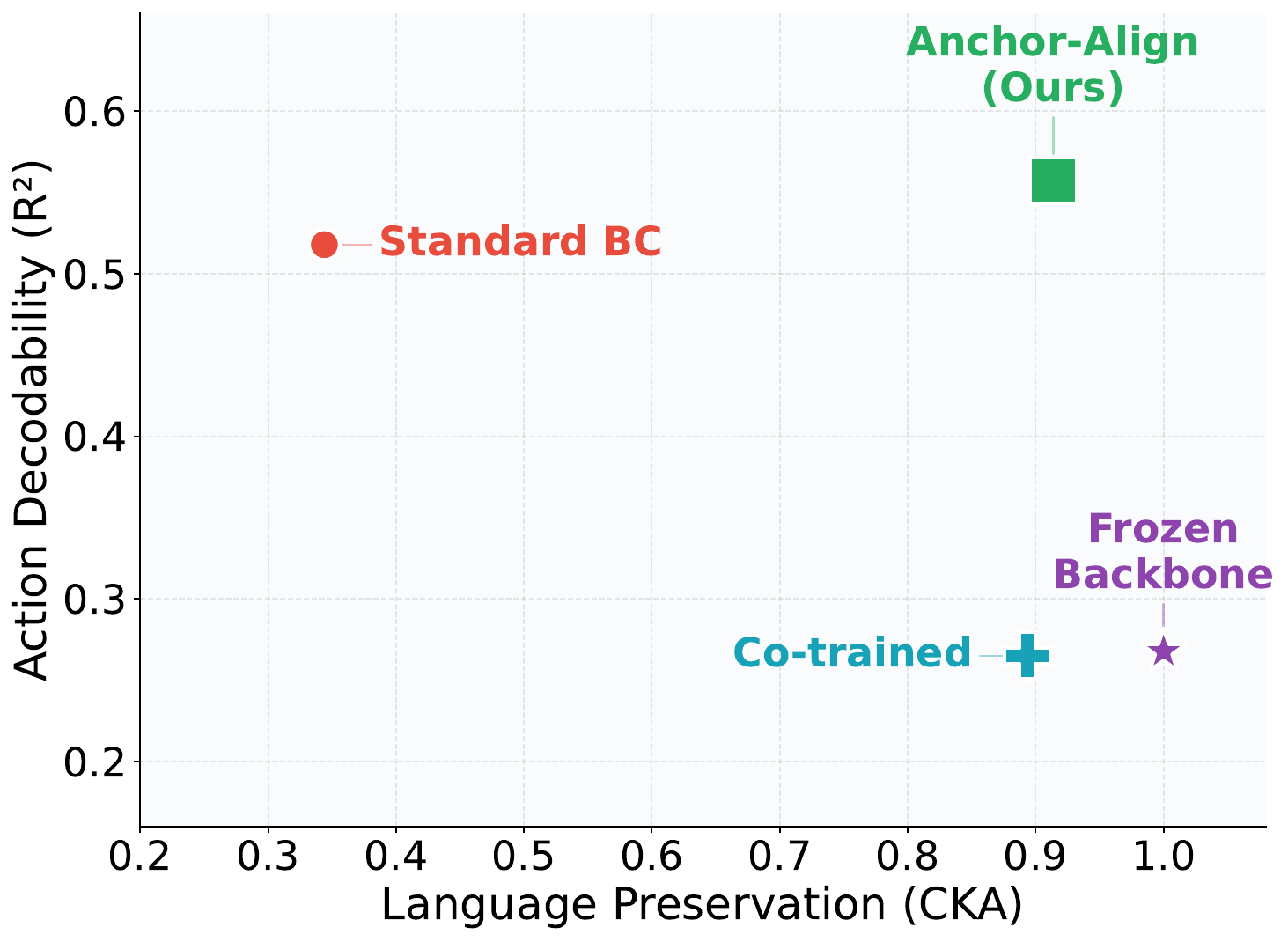}
        \caption{\small \textbf{Action decodability vs.\ language preservation.} Per-model language preservation (text-token CKA) vs.\ action decodability (linear-probe R$^2$): \ours attains the highest action decodability while sustaining high language preservation, gaining action information without overwriting the pretrained geometry.}
        \label{fig:preservation_decodability}
    \end{minipage}\hfill
    \begin{minipage}[t]{0.53\linewidth}
        \centering
        \includegraphics[width=\linewidth]{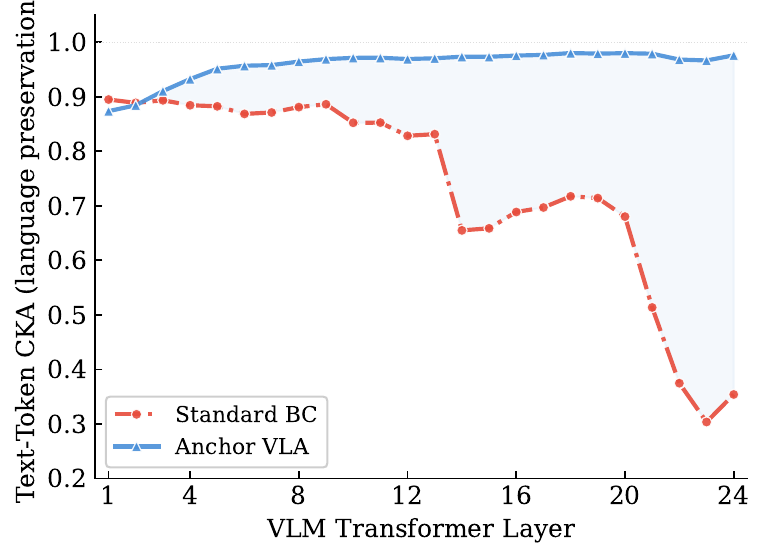}
        \caption{\small \textbf{Layer-wise language preservation.} Per-layer text-token CKA between the finetuned backbone and the pretrained VLM. Standard BC collapses the pretrained text representations in the output layers, whereas \anchor preserves them across all layers through layer-wise distillation.}
        \label{fig:language_preservation}
    \end{minipage}
\end{figure}

App.~\ref{subsec:preservation_decodability} reports language preservation per model; here we break the same measurement down by decoder layer, asking whether the GQA collapse in \cref{fig:catastrophic_forgetting} is only behavioral or whether finetuning actually reshapes the backbone's internal representations. \cref{fig:language_preservation} plots the text-token CKA between the finetuned backbone and the pretrained VLM at every layer (formal definition in App.~\ref{subsec:cka}). Standard BC progressively destroys the pretrained text representations with depth, collapsing in the output layers (CKA drops to 0.34 at layer 24), the same layers that drive the GQA collapse, so the degradation is representational rather than merely behavioral. \anchor instead recovers near-perfect preservation through layer-wise distillation, with an average CKA of 0.95 across layers.

\subsection{Directional information in backbone representation}
\label{app:direction_understanding}

To further unpack what \ourmethod changes inside the backbone, we linearly probe six motion-direction classes from the final-layer hidden state immediately before the action tokens. We compare the frozen backbone, \vlaadapter, Anchor-only VLA, and \ours using the same probing protocol: a separate linear classifier is fitted to each fixed backbone on 8{,}000 Spatial samples for five epochs and evaluated on 1{,}500 samples each from Spatial (in-distribution) and Object (out-of-distribution). Only the linear classifier is optimized during probing. In every case the probe reads the pre-action hidden state $\mathbf{h}^{\text{pre}}$ (Sec.~\ref{sec:lang-action-align}) directly from the backbone. For \ours, this bypasses the learned projection $\mathbf{W}_{\text{proj}}$ and the frozen language head used by the alignment loss, so the accuracies reflect directional information encoded in the VLM backbone itself rather than in the alignment head.

\cref{tab:direction_and_alignment} shows that Anchor-only VLA improves over the frozen backbone by 26.9 percentage points on Spatial and 5.7 points on Object. Relative to \vlaadapter, anchoring alone yields lower Spatial accuracy but higher Object accuracy (28.5\% vs.\ 21.4\%), consistent with preserving more transferable directional information. \ours achieves the highest accuracy on both suites. Together, these results show that \ours makes motion-direction information substantially more accessible in the VLM backbone, with the improvement over Anchor-only isolating the contribution of Language-Action Alignment (see \cref{tab:alignment_main}).

\begin{table}[h]
\centering
\caption{\textbf{Direction understanding.} Six-way linear-probe accuracy (\%) from fixed backbone representations. Anchor-only VLA improves over the frozen backbone on both suites and over \vlaadapter on Object OOD; \ours achieves the highest reported accuracy on both suites.}
\label{tab:direction_and_alignment}
\begin{tabular}{lcc}
\toprule
\textbf{Method} & \textbf{Spatial (in-dist.)} & \textbf{Object (OOD)} \\
\midrule
Frozen \vlaadapter{} backbone & 43.8 & 22.8 \\
\vlaadapter & 86.0 & 21.4 \\
Anchor-only VLA & 70.7 & 28.5 \\
\ours & \textbf{94.4} & \textbf{30.3} \\
\bottomrule
\end{tabular}
\end{table}

\section{Language-Action Alignment Diagnostic}
\label{sec:diagnostic}

\subsection{Diagnostic Protocol}
\label{benchmark_curation}

Sec.~\ref{sec:lang-action-align} defines the training-time direction labels and alignment loss, and \appref{subsec:align_head} specifies the alignment head. Here we describe the evaluation protocol for measuring whether the language and action heads agree on motion direction during robot rollouts.

\noindent\textbf{Per-frame action labels.}
For a trajectory frame $t$, we extract the world-coordinate translational displacement $\mathbf{d}_t=(\Delta x_t,\Delta y_t,\Delta z_t)$ from the policy action and apply the dominant-axis rule in Sec.~\ref{sec:lang-action-align}. The axis pairs map to forward/backward ($x$), left/right ($y$), and up/down ($z$), with ties resolved in the order $x$, $y$, $z$. This produces an action-direction label $\ell_t^{\mathrm{act}}$ for each frame.

\noindent\textbf{Diagnostic frame selection.}
We retain a frame only if $\|\mathbf{d}_t\|_2\geq\mu$, with $\mu=0.001$, and its direction label agrees with those of the next five frames: $\ell_t^{\mathrm{act}}=\ell_{t+j}^{\mathrm{act}}$ for $j\in\{1,\ldots,5\}$. These criteria remove near-stationary and transient motions from the diagnostic. They are distinct from the training-time filter on chunk-averaged translations, whose threshold $\tau$ is specified in \appref{subsec:align_head}.

\noindent\textbf{Language-action agreement.}
Given VLMs' ability to reason about spatial relations in visual question answering~\citep{Chen_2024_CVPR,Song_2025_CVPR,dalal-etal-2026-compositional,ICLR2026_288b63aa}, we test whether the backbone VLM's answers to motion-direction questions agree with the finetuned VLA's predicted actions.
On the same evaluated observation, we query the language head with a
templated question about the robot's immediate next motion:
\textit{``Q: In which immediate direction should the end effector move
to place the bowl on the plate?
Options: (a) Left, (b) Right, (c) Up, (d) Down,
(e) Forward, (f) Backward.''} The language prediction is interpreted in the camera-centric coordinate frame, i.e., left/right, up/down, and forward/backward denote the corresponding directions relative to the camera view. Because the policy action is expressed in the world/action coordinate frame, we convert the predicted camera-frame direction to the world frame using the known camera-to-world coordinate mapping. We then apply the same dominant-axis/sign convention used for the action displacement, yielding a world-frame language-direction label $\ell_t^{\mathrm{lang}}$. Language-action agreement is measured as $\mathbbm{1}[\ell_t^{\mathrm{lang}}=\ell_t^{\mathrm{act}}]$. This evaluation query is separate from the training loss, which supervises direction logits from the pre-action hidden state (Sec.~\ref{sec:lang-action-align}). Alignment is the fraction of evaluated observations for which the converted language prediction agrees with the action-direction label, regardless of whether the shared prediction is correct. We additionally compute alignment per rollout and correlate it with binary task success using Pearson $r$.

\clearpage
\subsection{Relationship to Task Success}
\label{reliability_of_language}

\begin{table}[htbp]
\centering
\setlength{\tabcolsep}{4pt}
\renewcommand{\arraystretch}{1.15}
\small
\caption{\textbf{Language-action alignment on LIBERO-PRO rollouts.} Success and alignment are percentages; Pearson $r$ correlates per-rollout alignment with task success.}
\label{tab:alignment_main}
\begin{adjustbox}{max width=\linewidth}
\begin{tabular}{l c >{\columncolor{LightBlue}}c c}
\toprule
VLA Model & Success & Alignment & Pearson $r$\\
\midrule
Co-training + KI~\citep{driess2025knowledge} & 48.5 & 14.3 & $+0.10$ \\
\vlaadapter~\citep{wang2025vla} & 61.0 & 16.8 & $-0.03$ \\
\molmoact~\citep{lee2025molmoact} & 53.5 & 13.2 & $-0.18$ \\
\ours & \textbf{71.9} & \textbf{78.4} & $\mathbf{+0.51}$ \\
\bottomrule
\end{tabular}
\end{adjustbox}
\end{table}

\cref{tab:alignment_main} applies this diagnostic to LIBERO-PRO rollouts and reports task success alongside the agreement metrics defined above. Co-training + KI is our \vlaadapter-based reimplementation described in \appref{app:baseline_details}.

\vlaadapter and Co-training + KI show low alignment ($16.8\%$ and $14.3\%$) and weak correlations with task success ($r=-0.03$ and $+0.10$). MolmoAct likewise has low alignment ($13.2\%$), with a negative correlation ($r=-0.18$). \ours raises alignment to $78.4\%$ and yields a positive correlation of $r=+0.51$, alongside $71.9\%$ task success compared with $61.0\%$ for \vlaadapter. Thus, higher agreement is associated with successful rollouts for \ours, while agreement alone is not a reliable success indicator across all evaluated methods. This correlation does not establish a causal effect of alignment on task success.

\section{Real-World Rollouts}
\label{real_world_rollouts}

\subsection{Bimanual Height Coordination}
\label{subsec:bimanual_height_coordination}

We compare vertical coordination on Tomato Plate Transfer between \baseline and \ourmethod using the same dual-arm YAM Ultra setup, parallel arm bases, and plate placement (App.~\ref{subsec:realworld_impl}). Gripper heights are reconstructed from recorded joint states using forward kinematics and smoothed with a centered nine-frame cubic filter before extracting the analysis interval.

We analyze 20 trials per method over the first 30 samples at 30\,Hz after $t_0$ ($0$--$0.967$\,s), where $t_0$ is the later of the two grippers' first sustained closures from an open state (below $0.2$ for three consecutive samples).

\begin{figure}[H]
    \centering
    \includegraphics[width=\linewidth]{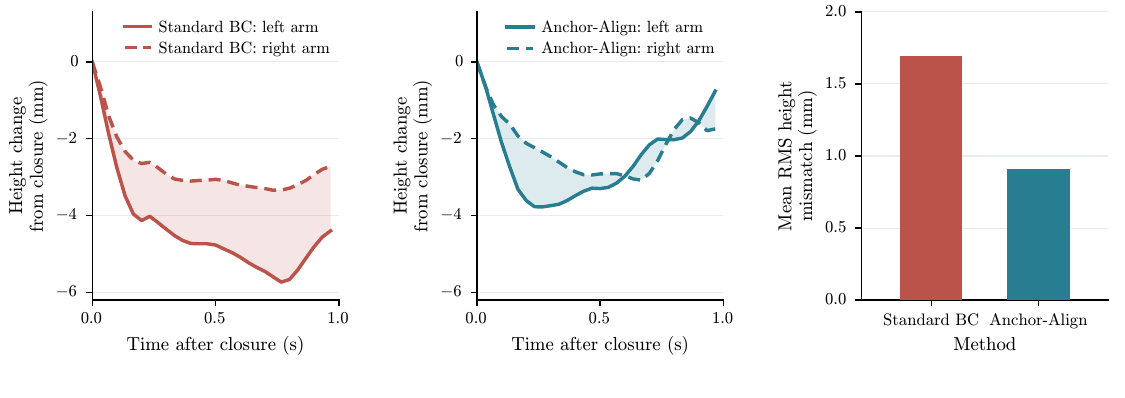}
    \caption{\small \textbf{Bimanual height coordination after gripper closure.} Left and middle: example Standard BC (red) and \ourmethod (teal) trajectories; solid and dashed curves show the left and right grippers' height changes from closure, with shading marking their difference. Right: mean per-trial RMS height-change mismatch across all eligible trials, showing lower mismatch with \ourmethod.}
    \label{fig:plate_height_alignment}
\end{figure}

Let $\Delta z_a(t)=z_a(t)-z_a(t_0)$ be the height change of arm $a\in\{L,R\}$ from closure. We measure each trial's root-mean-square (RMS) height-change mismatch as
\begin{equation}
    E_z = \sqrt{\frac{1}{N}\sum_{k=0}^{N-1}
    \bigl[\Delta z_R(t_k)-\Delta z_L(t_k)\bigr]^2},
    \qquad N=30,\quad t_k=t_0+\frac{k}{30}\,\mathrm{s}.
    \label{eq:bimanual_height_mismatch}
\end{equation}
The mean of $E_z$ across trials decreases from $1.69$\,mm for Standard BC to $0.91$\,mm for \ourmethod, a $46.2\%$ reduction (\cref{fig:plate_height_alignment}). The lower mismatch provides kinematic evidence of more coordinated vertical motion during the initial post-closure interval. This metric measures changes in relative gripper height; it does not directly establish plate orientation or isolate the contribution of either training loss.

\subsection{Object-Orientation Perturbation}
\label{subsec:object_orientation}

\begin{figure}[H]
    \centering
    \includegraphics[width=0.8\linewidth]{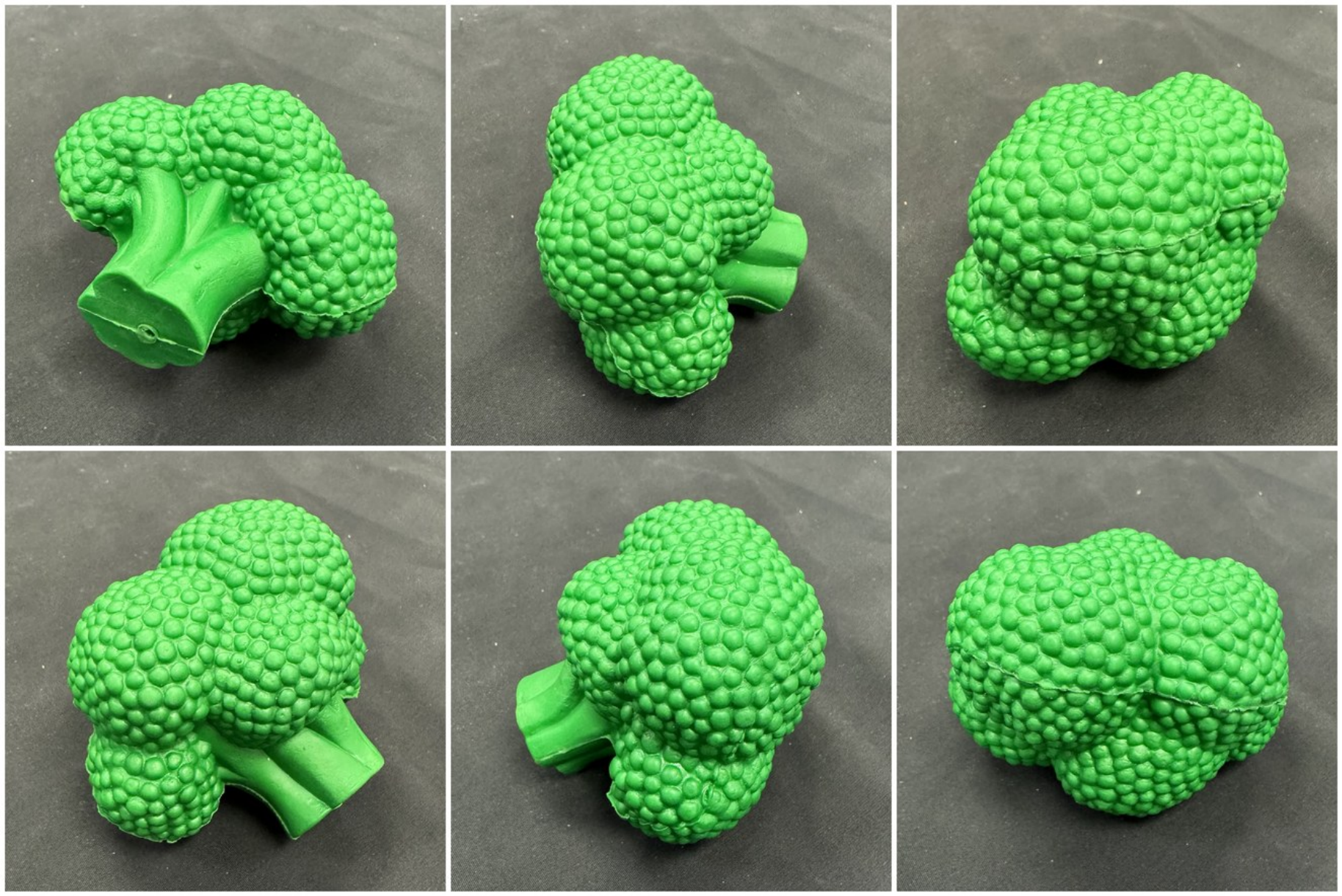}
    \caption{\small \textbf{Target-object orientation perturbation in Broccoli Placement.} The same broccoli, cropped from the front-view camera, shown in six distinct orientations used across evaluation rollouts of the "Pick up the green broccoli and place it on the plate task". We perturb the orientation of the object \emph{itself}, not just its position, so its silhouette and visible structure change substantially while its identity and the language instruction remain fixed. This prevents the policy from relying on a canonical, memorized view of the target and, combined with the position and distractor swaps in \cref{fig:ours_broccoli_rollouts}, makes the task considerably harder than the position-only swap of LIBERO-PRO.}
    \label{fig:broccoli_orientations}
\end{figure}

Beyond swapping object positions, our Broccoli Placement task also perturbs the \emph{orientation} of the target object itself. \cref{fig:broccoli_orientations} shows the broccoli used in the pick-and-place task in six distinct poses: varying the object's orientation while holding its identity and the instruction fixed prevents the policy from matching a memorized canonical view and forces it to re-ground perception in each rollout.

\subsection{Anchor-Align Real-World Rollouts}
\label{subsec:anchor_align_rollouts}

\begin{figure}[H]
    \centering
    \includegraphics[width=\linewidth]{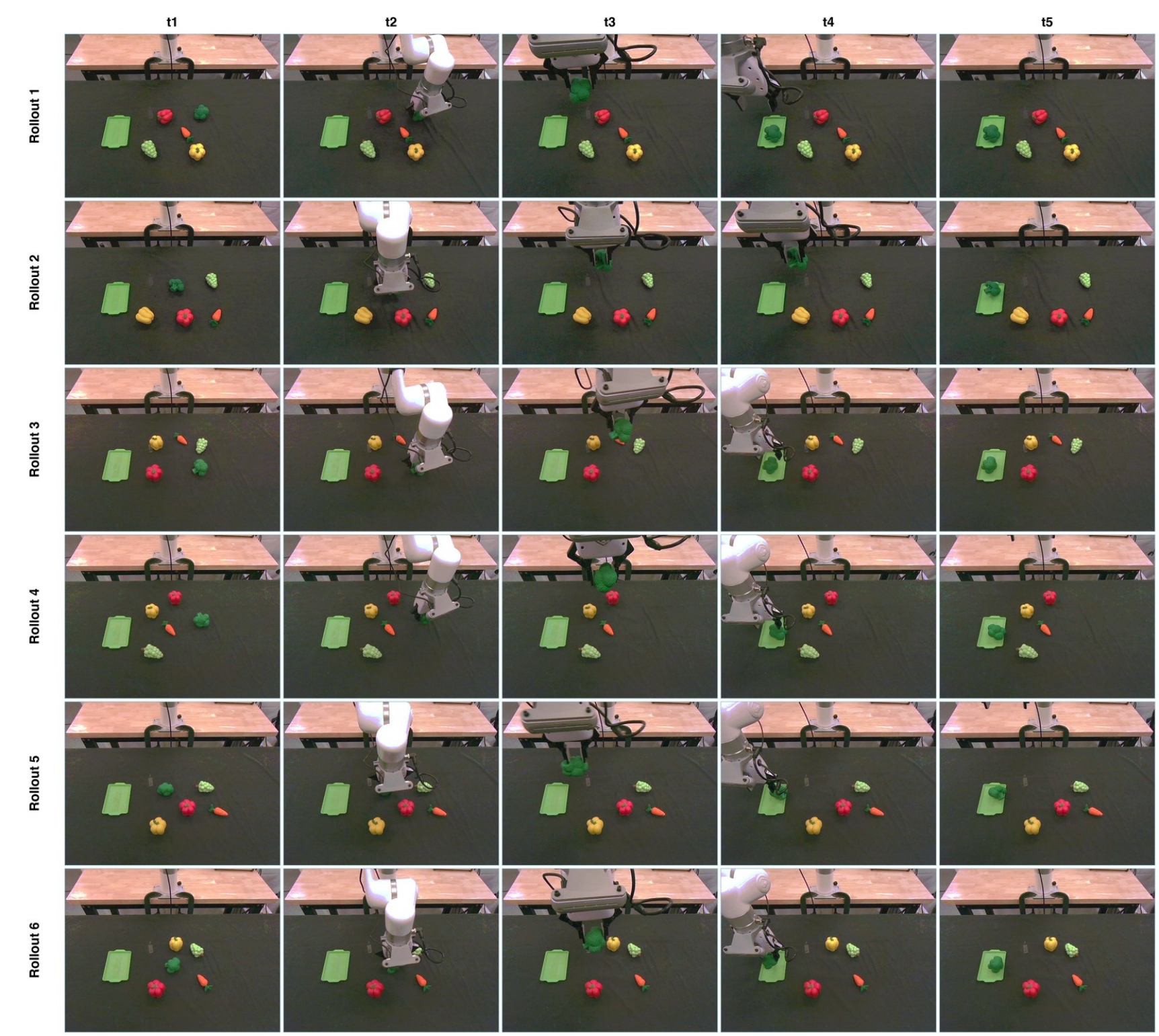}
    \caption{\small \textbf{Broccoli Placement rollouts from the \ours model.} The task is to pick up the broccoli and place it on the green plate.\textbf{ The broccoli's spatial position and the surrounding distractor objects are simultaneously swapped across rollouts}, so each rollout is evaluated on a unique scene configuration. All six rollouts succeed, showing that \ours handles joint object-swap and position-swap perturbations robustly without relying on memorized scene configurations. Each row shows 5 keyframes from a single rollout.}
    \label{fig:ours_broccoli_rollouts}
\end{figure}

\newpage

\begin{figure}[H]
    \centering
    \includegraphics[width=\linewidth]{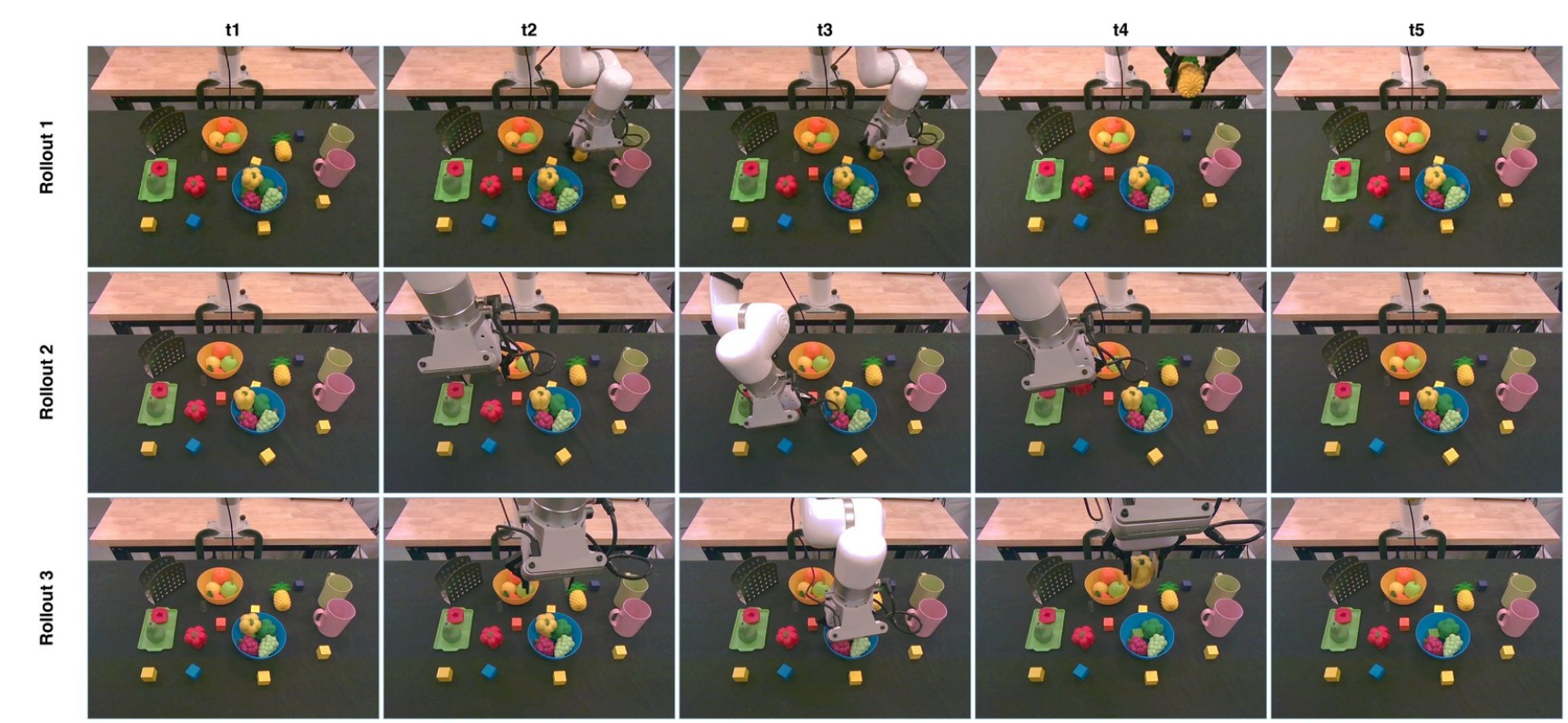}
    \caption{\small \textbf{Real-world rollouts from the \ours model.} The task is to pick up the object named by the language instruction (pineapple, red bell pepper, or yellow bell pepper) and place it on the plate in a \textbf{very cluttered scene}.\textbf{ The target is specified by text while many distractor objects (including the other candidate items) crowd the workspace}, so the policy must ground the instruction in the current observation and grasp the correct object rather than defaulting to a memorized choice. All three rollouts succeed, showing that \ours discriminates the language-specified target under heavy clutter and visually similar distractors. Rollouts 1, 2, and 3 target the pineapple, the red bell pepper, and the yellow bell pepper, respectively. Each row shows 5 keyframes from a single rollout.}
    \label{fig:ours_clutter_rollouts}
\end{figure}

\begin{figure}[H]
    \centering
    \includegraphics[width=\linewidth]{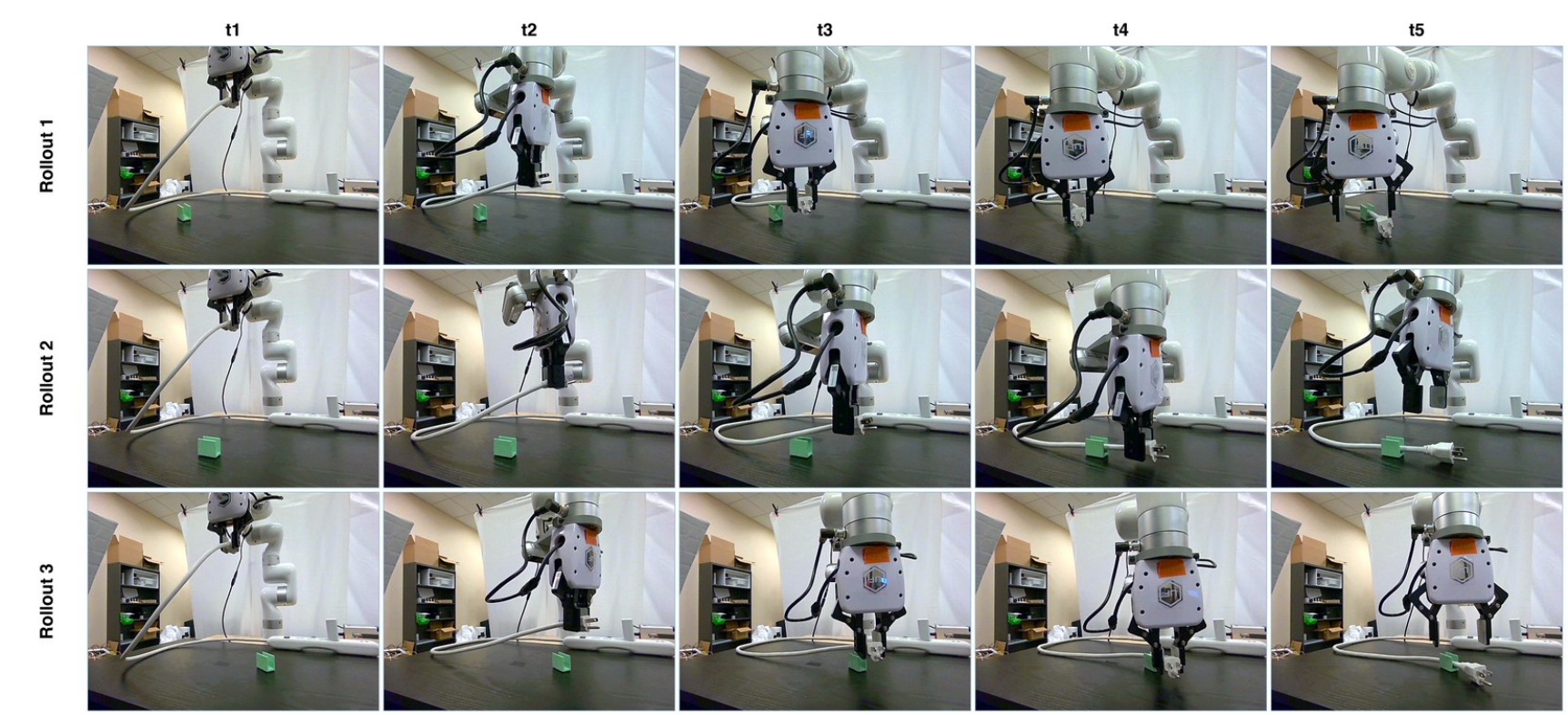}
    \caption{\small \textbf{Cable Insertion rollouts from the \ours model.} The task is to insert the wire into the green cable manager, an OOD target that replaces the red cable manager used during training. \textbf{Each row is a successful rollout in a different spatial configuration of the cable manager.} Each row shows 5 evenly spaced keyframes from a single rollout.}
    \label{fig:ours_cable_rollouts}
\end{figure}

\newpage

\begin{figure}[H]
    \centering
    \includegraphics[width=\linewidth]{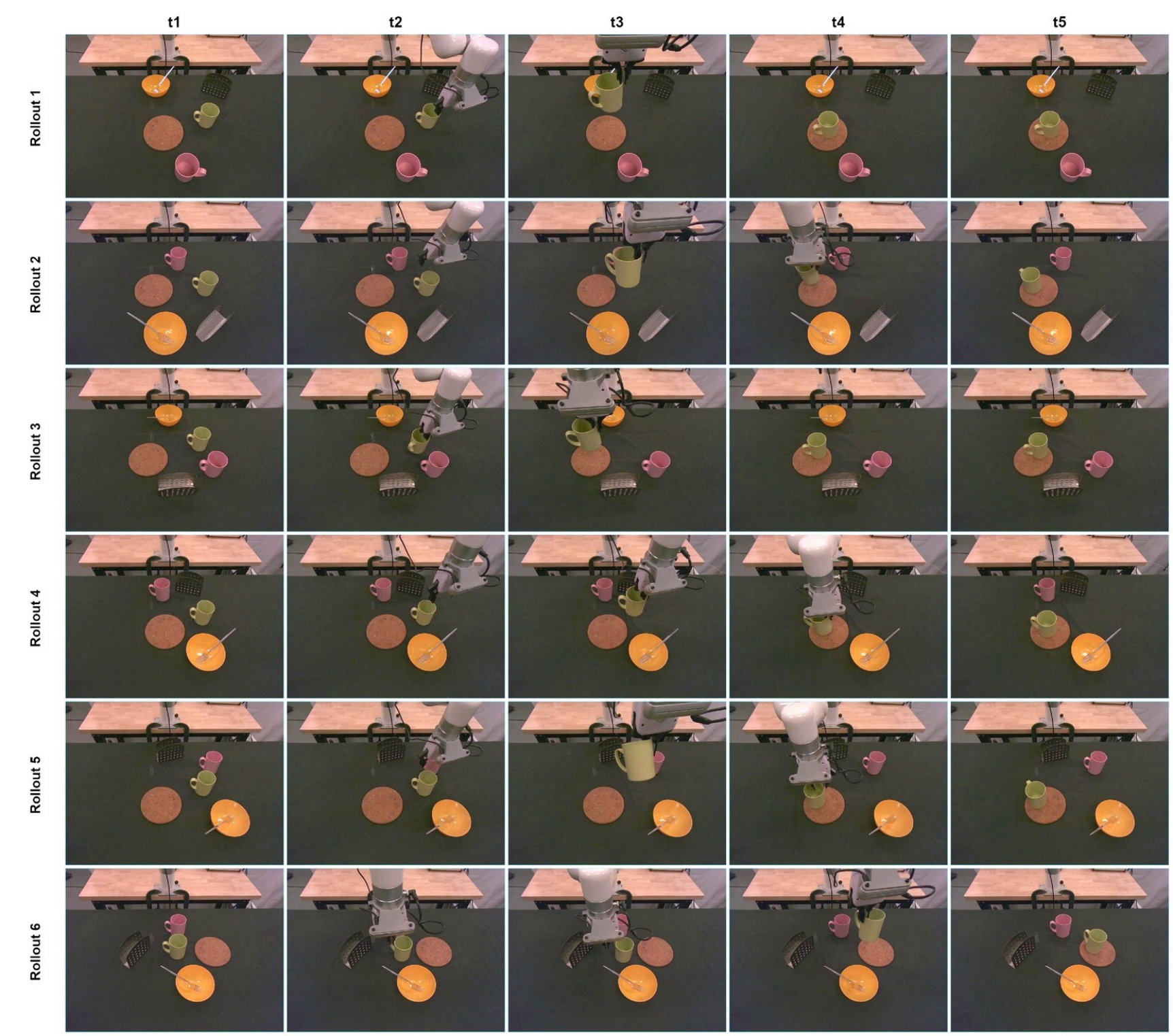}
    \caption{\small \textbf{Green Mug Placement rollouts from the \ours model.} The task is to pick up the green mug and place it on the plate.\textbf{ The mug's spatial position and the surrounding distractor objects are simultaneously swapped across rollouts}, so each rollout is evaluated on a unique scene configuration. All six rollouts succeed, showing that \ours handles joint object-swap and position-swap perturbations robustly without relying on memorized scene configurations. Each row shows 5 evenly spaced keyframes from a single rollout.}
    \label{fig:ours_mug_rollouts}
\end{figure}

\begin{figure}[H]
    \centering
    \includegraphics[width=\linewidth]{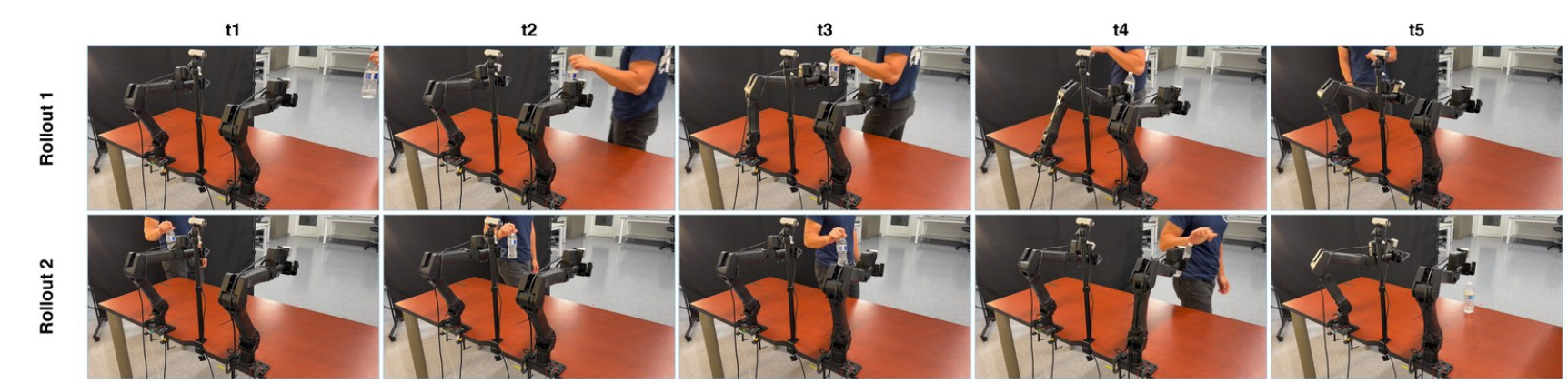}
    \caption{\small \textbf{Reactive Grasp rollouts from the \ours model.} A person carries a bottle across the robot's workspace, and the robot grasps the bottle while it moves. The person walks in opposite directions in the two rollouts, so a different arm makes the grasp. Each row shows 5 keyframes from a single rollout.}
    \label{fig:ours_reactive_grasp_rollouts}
\end{figure}

\section{Qualitative Results in LIBERO Simulator}
\label{app:libero_rollouts}

% We present a qualitative comparison of policy rollouts between standard BC and \ours on: 1) LIBERO-PRO, which tests generalization to semantic perturbations (e.g., object swap, language rephrase), and 2) LIBERO Plus, which tests robustness to environmental perturbations (e.g., lighting, viewpoint, background texture). Across all figures (\cref{fig:goal_plus_rollouts_1,fig:goal_plus_rollouts_2,fig:l10_plus_rollouts_1,fig:l10_plus_rollouts_2,fig:libero_pro_object_rollouts}), standard BC fails under these distribution shifts, while Anchor-Align completes the task successfully. Additional qualitative rollouts are available on our project page: \url{https://anchorvla2026.github.io/AnchorVla2026/}. In these figures, the baseline \baseline is labeled \emph{Action-Only} and \ours is labeled \emph{Anchor-Align (Ours)}.

\begin{figure}[H]
    \centering
    \includegraphics[width=0.9\linewidth]{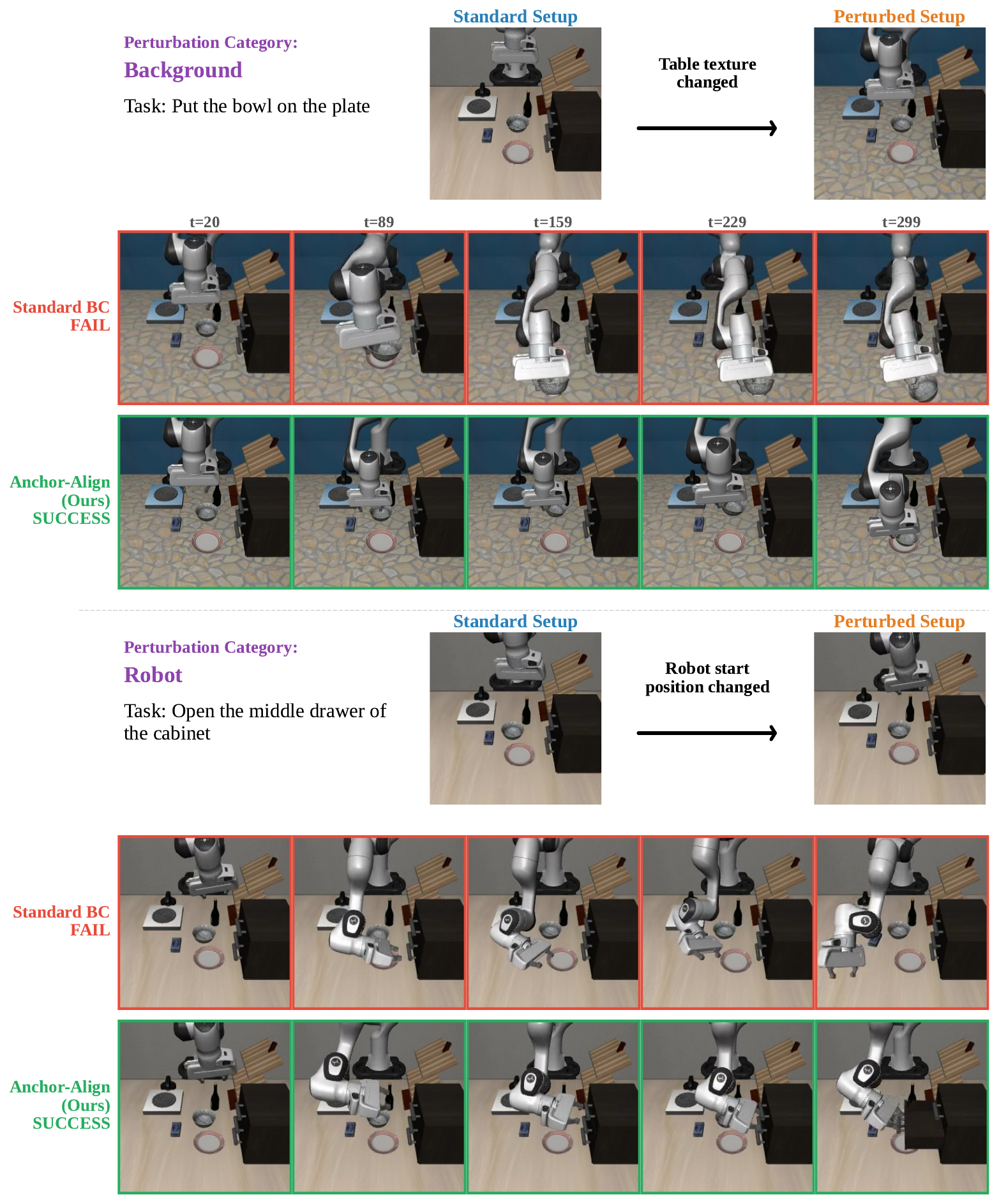}
    \caption{\small \textbf{LIBERO-Goal Plus: \emph{Background} and \emph{Robot} perturbations.} \textit{Top:} Background perturbation on ``Put the bowl on the plate''---the table texture is changed to stone tile. \baseline (red) collapses onto the table without releasing the bowl on the plate; \ours (green) places the bowl on the plate. \textit{Bottom:} Robot perturbation on ``Open the middle drawer of the cabinet''---the robot's initial joint configuration is shifted. \baseline approaches the wrong side of the cabinet and never opens the drawer; \ours locates the middle drawer and pulls it open.}
    \label{fig:goal_plus_rollouts_1}
\end{figure}

\newpage

\begin{figure}[H]
    \centering
    \includegraphics[width=\linewidth]{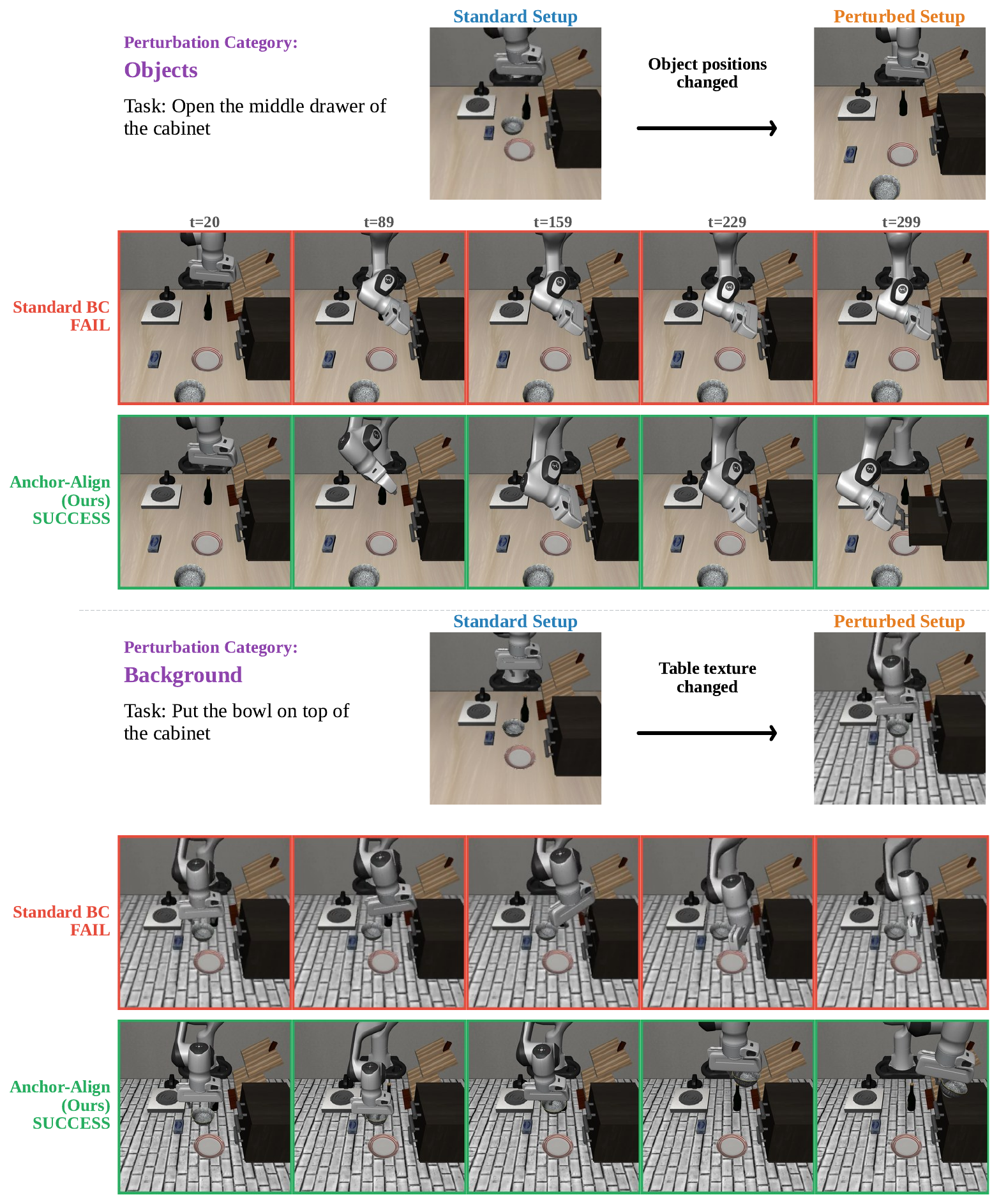}
    \caption{\small \textbf{LIBERO-Goal Plus: \emph{Objects} and \emph{Background} perturbations.} \textit{Top:} Objects perturbation on ``Open the middle drawer of the cabinet''---the cabinet, plate, and surrounding distractors are rearranged. \baseline (red) hovers above the wrong region without contacting the handle; \ours (green) locates the middle drawer and opens it. \textit{Bottom:} Background perturbation on ``Put the bowl on top of the cabinet''---the wood floor is swapped for a stone-tile surface. \baseline drifts away from the cabinet; \ours grasps the bowl and deposits it on top of the cabinet.}
    \label{fig:goal_plus_rollouts_2}
\end{figure}

\begin{figure}[H]
    \centering
    \includegraphics[width=\linewidth]{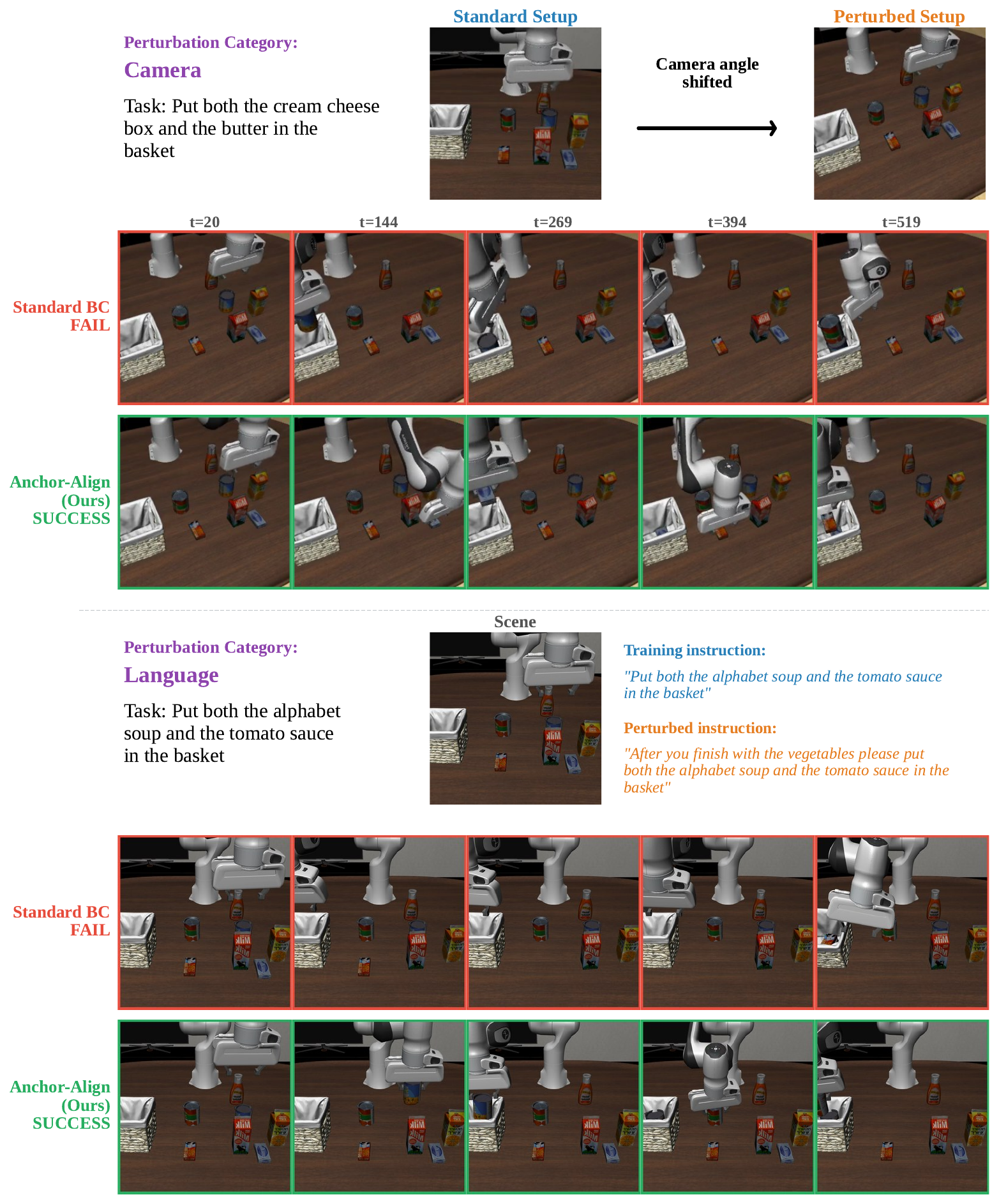}
    % \caption{\small \textbf{LIBERO-Long Plus: \emph{Camera} and \emph{Language} perturbations.} \textit{Top:} Camera perturbation on the long-horizon task ``Put both the cream cheese box and the butter in the basket''---the camera viewpoint is shifted. \baseline (red) picks up the incorrect object and places it in the basket showcasing ; \ours (green) sequentially picks both targets and drops them in the basket. \textit{Bottom:} Language perturbation on ``Put both the alphabet soup and the tomato sauce in the basket''---the instruction is rephrased to ``\emph{After you finish with the vegetables please put both the alphabet soup and the tomato sauce in the basket}''. \baseline fails to initiate the correct grasps; \ours follows the rephrased instruction to completion.}
    \caption{\small \textbf{LIBERO-Long Plus: \emph{Camera} and \emph{Language} perturbations.} \textit{Top:} Camera perturbation on the long-horizon task ``Put both the cream cheese box and the butter in the basket'': the camera viewpoint is shifted. \baseline (red) picks up the incorrect object and places it in the basket, showcasing its reliance on memorized scene-to-action mappings that break under the viewpoint shift; \ours (green) sequentially picks both targets and drops them in the basket. \textit{Bottom:} Language perturbation on ``Put both the alphabet soup and the tomato sauce in the basket'': the instruction is rephrased to ``\emph{After you finish with the vegetables please put both the alphabet soup and the tomato sauce in the basket}''. \baseline fails to initiate the correct grasps; \ours follows the rephrased instruction to completion.}
    \label{fig:l10_plus_rollouts_1}
\end{figure}

\begin{figure}[H]
    \centering
    \includegraphics[width=\linewidth]{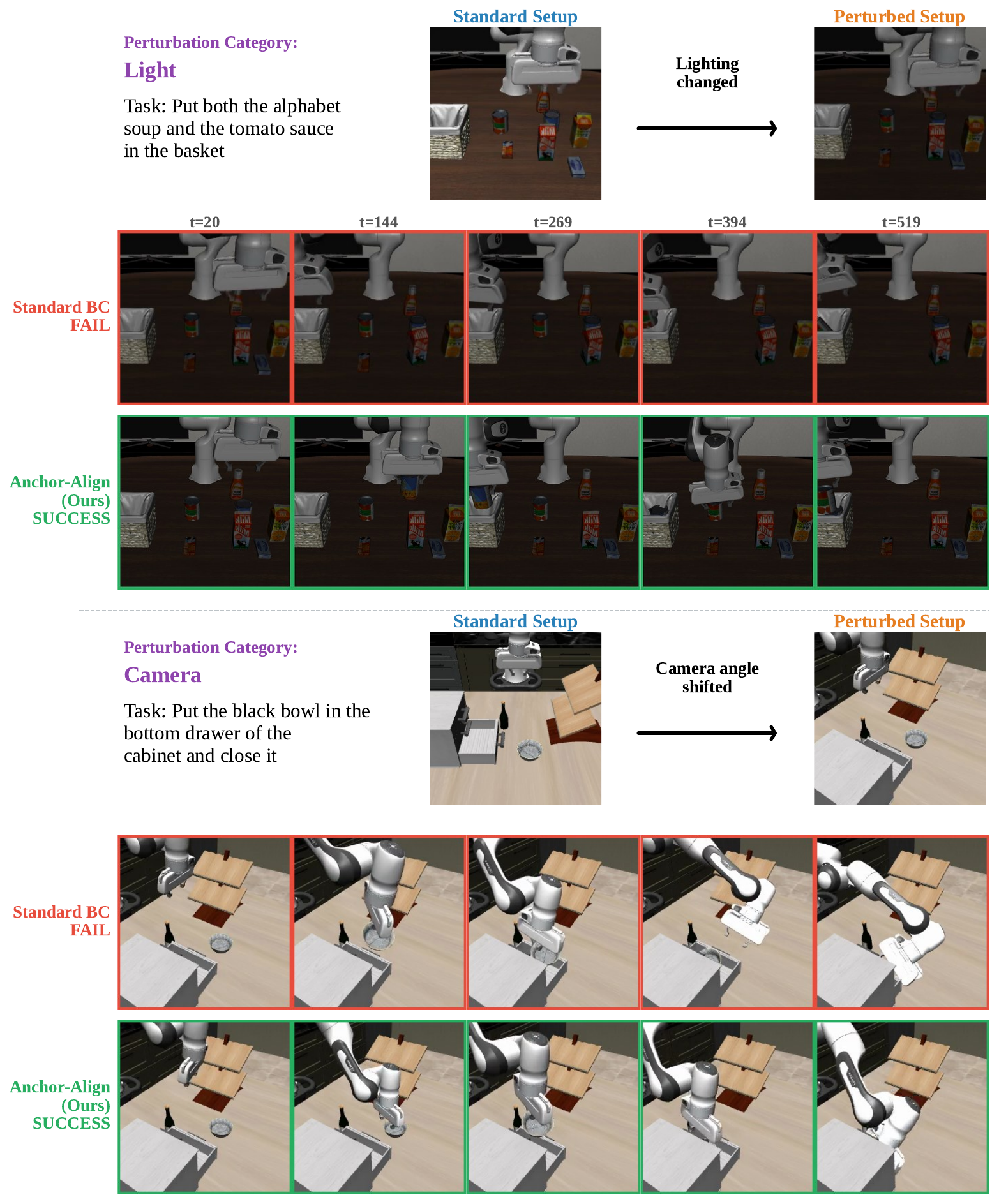}
    \caption{\small \textbf{LIBERO-Long Plus: \emph{Lighting} and \emph{Camera} perturbations.} \textit{Top:} Lighting perturbation on ``Put both the alphabet soup and the tomato sauce in the basket'': the scene is darkened. \baseline (red) grasps an incorrect object, places it in the basket, and then idles without ever reaching for the second target; \ours (green) deposits both items in the basket. \textit{Bottom:} Camera perturbation on the long-horizon task ``Put the black bowl in the bottom drawer of the cabinet and close it'': the camera viewpoint is shifted. \baseline pulls the wrong drawer and never completes the close step; \ours opens the bottom drawer, places the black bowl inside, and closes it.}
    \label{fig:l10_plus_rollouts_2}
\end{figure}

\begin{figure}[H]
    \centering
    \includegraphics[width=\linewidth]{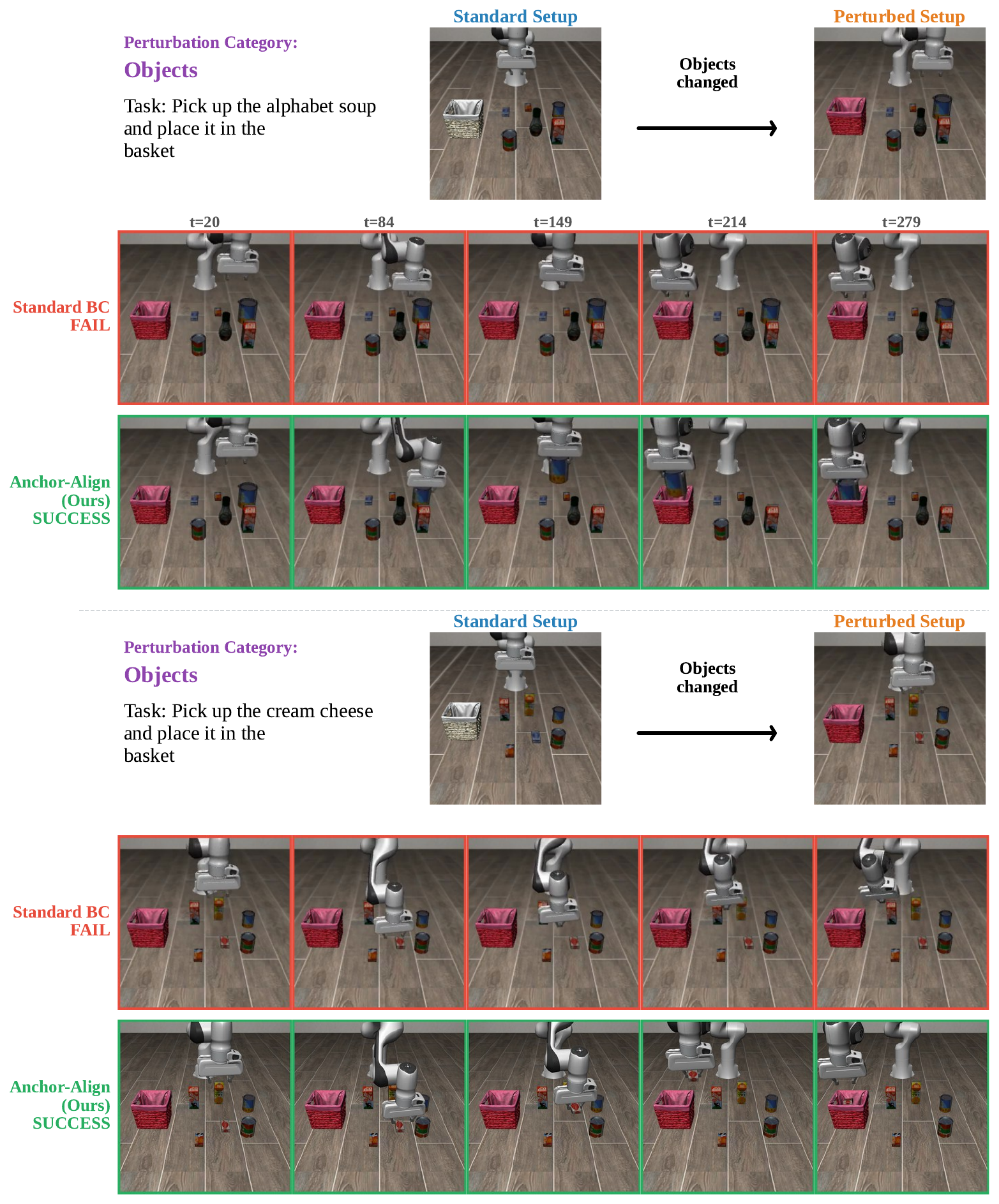}
    \caption{\small \textbf{LIBERO-PRO \emph{Object Swap}.} Both rows use the same object-swap regime: the basket is replaced with a pink wicker variant and the surrounding canned items are exchanged for novel objects unseen during training. \textit{Top:} ``Pick up the alphabet soup and place it in the basket'': \baseline (red) hovers over an incorrect target and then moves to the basket without grasping any object, whereas \ours (green) grounds to the alphabet-soup can and deposits it in the basket. \textit{Bottom:} ``Pick up the cream cheese and place it in the basket'': \baseline correctly identifies the target object but fails to grasp it, completing its trajectory with an empty gripper, whereas \ours selects the cream-cheese box and places it correctly.}
    \label{fig:libero_pro_object_rollouts}
\end{figure}

\clearpage
\end{document}